%% file: main.tex
\documentclass[10pt,twocolumn,letterpaper]{article}

\usepackage[pagenumbers]{cvpr} 

\input{preamble}

\input{src/others/definition}

\usepackage{times}
\usepackage{epsfig}
\usepackage{pifont}
\usepackage{latexsym}
\usepackage{bbding}
\usepackage{multirow}
\usepackage{graphicx}
\usepackage{dashrule}
\usepackage{float}
\usepackage{rotfloat}

\usepackage[T1]{fontenc}

\usepackage[utf8]{inputenc}

\usepackage{microtype}
\usepackage{inconsolata}

\usepackage[utf8]{inputenc} 
\usepackage[T1]{fontenc}    
\usepackage{url}            
\usepackage{booktabs}       
\usepackage{amsfonts}       
\usepackage{nicefrac}       
\usepackage{microtype}      
\usepackage{xcolor}         
\usepackage{tabularx}

\usepackage{graphicx}
\usepackage{amsmath}
\usepackage{amssymb}
\usepackage{mdframed} 
\usepackage{mathptmx}
\usepackage{url}
\usepackage{wrapfig}
\usepackage{color}
\usepackage{colortbl}
\usepackage{svg}
\usepackage{makecell}
\usepackage{utfsym}
\usepackage{xspace}

\usepackage{graphicx}
\usepackage{enumitem}
\usepackage{multicol}
\usepackage{bbm}
\usepackage{calrsfs}
\usepackage{stmaryrd}
\usepackage{CJK}
\usepackage{dirtree}
\usepackage{pythonhighlight}
\usepackage{caption}
\usepackage{subcaption}
\usepackage[pdf]{graphviz}
\usepackage{arydshln} 

\usepackage{algorithm}
\usepackage{algorithmic}
\usepackage{bm}
\usepackage{mathtools}
\usepackage{lipsum}
\usepackage{cuted}
\usepackage{siunitx}

\definecolor{pearDark}{HTML}{2980B9}
\definecolor{blueColor}{HTML}{0054D6}
\definecolor{cyanColor}{HTML}{31E1C8}
\definecolor{purpleColor}{HTML}{A261FF}
\definecolor{yellowColor}{HTML}{FFDB13}

\definecolor{citecolor}{HTML}{0071BC}
\definecolor{linkcolor}{HTML}{0071BC}
\definecolor{pearDark}{HTML}{2980B9}

\newcommand{\sota}[1]{\cellcolor{pearDark!20}{#1}}
\newcommand{\figdir}{src/figures/}
\newcommand{\model}{GroupMixFormer}
\newcommand{\block}{GMA}

\usepackage[pagebackref=false,breaklinks=true,colorlinks,urlcolor=linkcolor,citecolor=citecolor,linkcolor=linkcolor,bookmarks=false]{hyperref}

\usepackage[capitalize,noabbrev]{cleveref}
\usepackage{appendix}
\usepackage{cleveref}
\crefname{section}{Section}{Sections}
\crefname{theorem}{Theorem}{Theorems}
\crefname{lemma}{Lemma}{Lemmas}
\crefname{equation}{Equation}{Equations}
\crefname{proposition}{Proposition}{Propositions}
\crefname{claim}{Claim}{Claims}
\crefname{appendix}{Appendix}{Appendices}
\crefname{algorithm}{Algorithm}{Algorithms}
\crefname{figure}{Figure}{Figs}
\crefname{table}{Table}{Tables}
\crefname{remark}{Remark}{Remarks}
\crefname{definition}{Definition}{Definitions}
\crefname{equation}{Equation}{Equations}
\crefname{corollary}{Corollary}{Corollaries}
\crefalias{section}{appendix}

\title{Advancing Vision Transformers with Group-Mix Attention}

\author{Chongjian Ge$^1$ \quad
  Xiaohan Ding$^2$ \quad
  Zhan Tong$^3$ \quad 
  Li Yuan$^4$ \quad 
  Jiangliu Wang$^5$ \quad
  Yibing Song$^6$ \quad
  Ping Luo$^1$\\
  {$^1$The University of Hong Kong} \quad {$^2$Tencent AI Lab} \quad {$^3$Ant Research}  \\
  {$^4$Peking University} \quad {$^5$The Chinese University of Hong Kong} \quad {$^6$AI${^3}$ Institute, Fudan University}  \\
 }

\begin{document}
\maketitle

\input{src/secs/0-abs}
\input{src/secs/1-intro}

\input{src/secs/2-related}
\input{src/secs/3-methods}

\input{src/secs/4-exp}

\input{src/secs/5-conclusion}

{
    \small
    \bibliographystyle{ieeenat_fullname}
    \bibliography{src/bibs/unsort}
}

\newpage
\quad
\newpage

\input{src/secs/6-appendix}

\end{document}

%% file: preamble.tex
\usepackage[dvipsnames]{xcolor}


%% file: src/others/definition.tex
\usepackage{xspace}
\usepackage{pifont}
\usepackage{booktabs,multirow}
\usepackage{tabularx}

\newlength\replength
\newcommand\repfrac{.66}

\newcommand\rulewidth{.8pt}
\newcommand\tdashfill[1][\repfrac]{\cleaders\hbox to \replength{%
  \smash{\rule[\arraystretch\ht\strutbox]{\repfrac\replength}{\rulewidth}}}\hfill}

\usepackage{diagbox}
\usepackage{etoc}
\etocdepthtag.toc{mtchapter}
\etocsettagdepth{mtchapter}{subsection}
\etocsettagdepth{mtappendix}{none}

\usepackage{mdframed}
\usepackage{pifont}
\usepackage{xcolor}
\definecolor{formalshade}{rgb}{0.95,0.95,1}
\definecolor{blueColor}{HTML}{0054D6}
\definecolor{cyanColor}{HTML}{31E1C8}

\usepackage{lipsum}
\newmdenv[
  skipabove=\topskip,
  skipbelow=\topskip,
  innermargin=0pt,
  outermargin=0pt,
  innerleftmargin=4pt,
  innerrightmargin=4pt,
  innertopmargin=2pt,
  innerbottommargin=2pt,
  topline=false,
  rightline=false,
  bottomline=false,
  linecolor=blueColor,
  linewidth=2pt,
   ]{tipbox*}

\newmdenv[
  skipabove=\topskip,
  skipbelow=\topskip,
  innermargin=0pt,
  outermargin=0pt,
  innerleftmargin=4pt,
  innerrightmargin=4pt,
  innertopmargin=2pt,
  innerbottommargin=2pt,
  topline=false,
  rightline=false,
  bottomline=false,
  linecolor=cyanColor,
  linewidth=2pt,
]{tipbox_a*}

%% file: src/secs/0-abs.tex
\begin{abstract}
Vision Transformers (ViTs) have been shown to enhance visual recognition through modeling long-range dependencies with multi-head self-attention (MHSA), which is typically formulated as Query-Key-Value computation. However, the attention map generated from the Query and Key  captures only token-to-token correlations at one single granularity. In this paper, we argue that self-attention should have a more comprehensive mechanism to capture correlations among tokens and groups (i.e., multiple adjacent tokens) for higher representational capacity. Thereby, we propose Group-Mix Attention (GMA) as an advanced replacement for traditional self-attention, which can simultaneously capture token-to-token, token-to-group, and group-to-group correlations with various group sizes. To this end, GMA splits the Query, Key, and Value into segments uniformly and performs different group aggregations to generate group proxies. The attention map is computed based on the mixtures of tokens and group proxies and used to re-combine the tokens and groups in Value. Based on GMA, we introduce a powerful backbone, namely GroupMixFormer, which achieves state-of-the-art performance in image classification, object detection, and semantic segmentation with fewer parameters than existing models. For instance, GroupMixFormer-L (with 70.3M parameters and $\mathit{384}^2$ input) attains 86.2\% Top-1 accuracy on ImageNet-1K without external data, while GroupMixFormer-B (with 45.8M parameters) attains 51.2\% mIoU on ADE20K.  Codes and trained models are released in \href{https://github.com/AILab-CVC/GroupMixFormer}{https://github.com/AILab-CVC/GroupMixFormer}.
\end{abstract}

%% file: src/secs/1-intro.tex
\section{Introduction} \label{sec:intro}

Vision Transformers (ViTs) significantly improve visual recognition tasks, including image classification~\citep{dosovitskiy-iclr21-vit,yuan2021volo}, self-supervised learning~\citep{chen-21iccv-mocov3,caron-21iccv-dino,xie-2021-swinself,bao-2021-beit}, object detection~\citep{liu-iccv21-swin,dai-21cvpr-dynamic}, and semantic segmentation~\citep{wang-iccv21-pvt,wang-21cvm-pvtv2,xie-21nips-segformer}. One crucial module that contributes significantly to the performance improvement is the multi-head self-attention (MHSA), which enables network designing with the long-range dependency modeling~\citep{vaswani-nips17-trans,raghu-2021-vision}, global receptive field, higher flexibility~\citep{jia-2021-clip,cordonnier-2019-relationship} and stronger robustness~\citep{paul-2021-robust,xie-21nips-segformer}. Typically, the term ``attention'' (i.e., the Q-K-V attention) means linearly re-combining \emph{Value} with the correlations between the \emph{Query} and \emph{Key}, which are usually computed between pairs of individual tokens.

\input{src/figure_latex/teaser-all}

\input{src/figure_latex/teaser-performance}

However, it's empirically found that there is a major limitation in Q-K-V self-attention, which is shown in ~\cref{fig:toy}: the attention map only describes the correlations between each individual token pairs at one single granularity (\cref{fig:toy}(a)), and multiplying the attention map with the Value only linearly re-combines the individual tokens. This framework obviously does not consider the correlations among different token groups (i.e., neighborhoods) at various granularities. For one specific example, self-attention does not correlate the nine tokens at the top-left corner as a whole to those groups at the bottom-right. This limitation, though obvious, has been unintentionally neglected because the Q-K-V computation seems to be capable enough of modeling the mappings from input to output, as any entry in the output attends to each individual entry in the input.

In this study, we propose a more comprehensive modeling approach, referred to as \emph{Group-Mix Attention} (\textbf{GMA}), to alleviate the aforementioned limitations of the widely used Q-K-V self-attention mechanism. \block\ splits the tokens into uniform and distinct segments and substitutes some individual tokens with group proxies generated via group aggregators, as shown in ~\cref{fig:toy} (b). Afterward, we compute the attention map with the Query and Key (where some tokens have been replaced by group proxies) and use it to re-combine both the group proxies together with individual tokens in Value. The proposed \block\ has some appealing advantages: \textbf{(1)} \block\ is capable of modeling correlations among not only individual tokens but also groups of tokens. Different kinds of attentions are \emph{mixed} to obtain a better understanding of the tokens from a comprehensive aspect. The token-to-token, token-to-group, and group-to-group correlations are simultaneously modeled \textit{within each single layer} for higher representational capabilities. \textbf{(2)} \block\ is efficient and easy to implement. The group-to-group correlation is computed via aggregating the groups into proxy tokens and then computing the correlation between proxies (as shown in ~\cref{fig:pipeline}). Such a process can be efficiently implemented with sliding-window-based operations, e.g., pooling and convolution.

Building on \block , we develop a hierarchical vision transformer, \model, which can serve as visual backbones for various tasks. We evaluate \model s on standard visual recognition tasks, including image classification, object detection, and semantic segmentation, and performed comparisons with advanced models as shown in \cref{fig:teaser_performance}. The results demonstrate the effectiveness of our designs. For example, a small \model\ instance (with 22.4M parameters) achieves 83.4\% Top-1 accuracy on ImageNet-1K, comparable to the much larger Swin-B\citep{liu-iccv21-swin} (88M parameters). 
Additionally, \model\ also performs favorably against state-of-the-art ViTs and CNNs on object detection and semantic segmentation. On the ADE20K dataset, \model-B achieves 51.2\% mIoU with a backbone size of 46M. Extensive experiments also demonstrate that effectively modeling the correlations among tokens and diverse groups is crucial for the success of \block. Such a design paradigm can also be readily adopted into other ViT architectures as an advanced replacement for traditional self-attention.

%% file: src/figure_latex/teaser-all.tex
\begin{figure}[t]
		\centering
		\begin{tabular}{c}
			\includegraphics[width=0.96\linewidth]{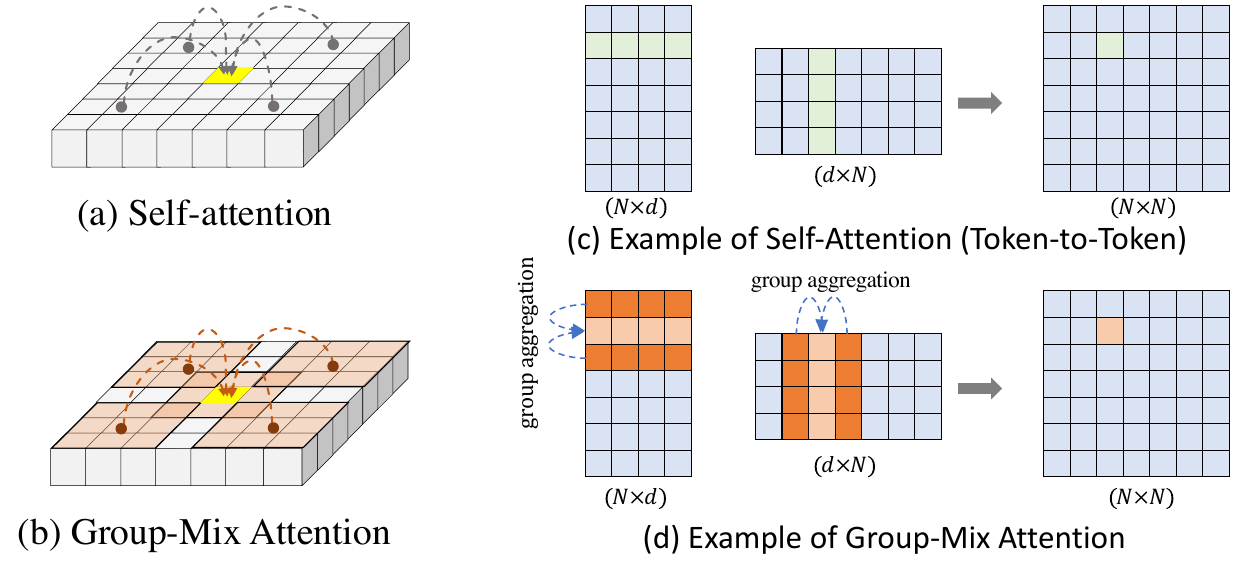}
		\end{tabular}
		\vspace{-0.5mm}
\caption{\textbf{Conceptual comparisons between the self-attention and our proposed Group-Mix Attention (GMA).} In (a) and (b), we showcase with 7$\times$7 single-dimensional tokens. 
Unlike the self-attention that computes correlations between pairs of individual tokens, \block\ creates proxies of token groups (e.g., nine adjacent tokens) via group aggregators, and then computes the group-to-group correlations via proxies. In (c) and (d), we show the concrete computation of \block\ with seven four-dimensional tokens, so that $N$=7 and $d$=4. To compute the correlations between two highlighted groups that each consist of three tokens, we aggregate them into two proxies for further multiplication. The group aggregation can be effectively implemented via sliding-window-based operators.}
\label{fig:toy}
\vspace{-1.5em}
\end{figure}

%% file: src/figure_latex/teaser-performance.tex
\def\swfive{0.31\linewidth}
\begin{figure*}[t]
		\begin{center}
			\begin{tabular}{ccc}
				\includegraphics[width=\swfive]{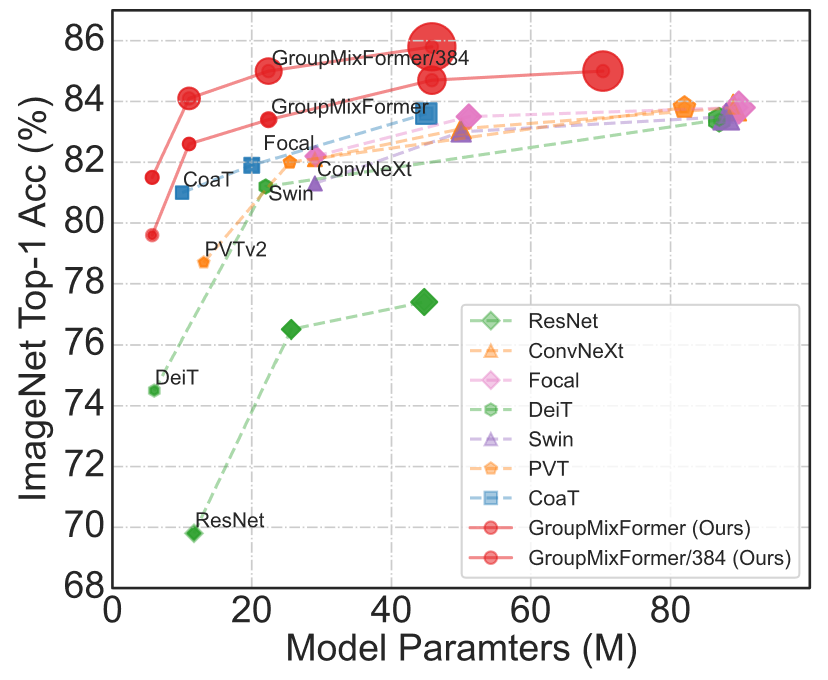}&
				\includegraphics[width=\swfive]{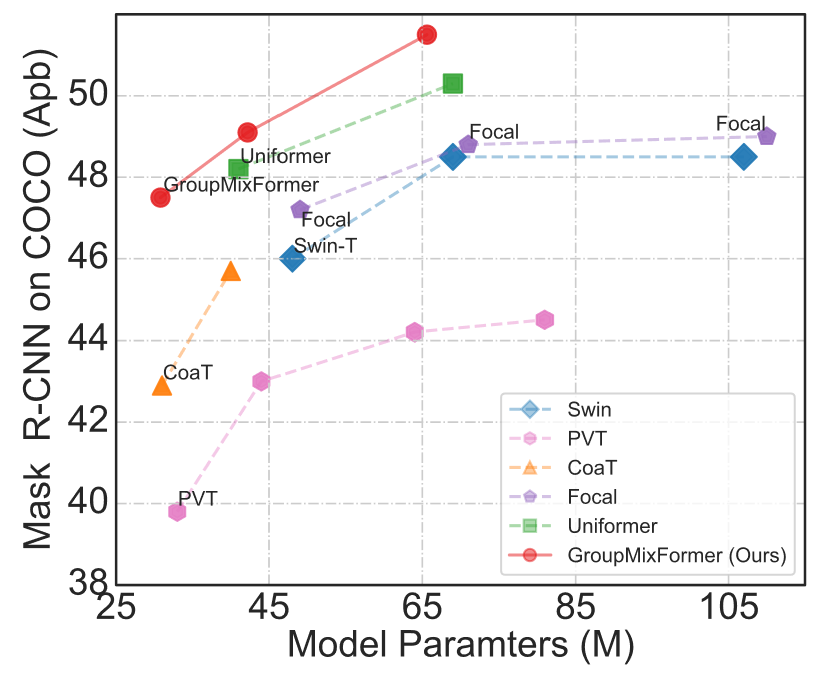} &
				\includegraphics[width=\swfive]{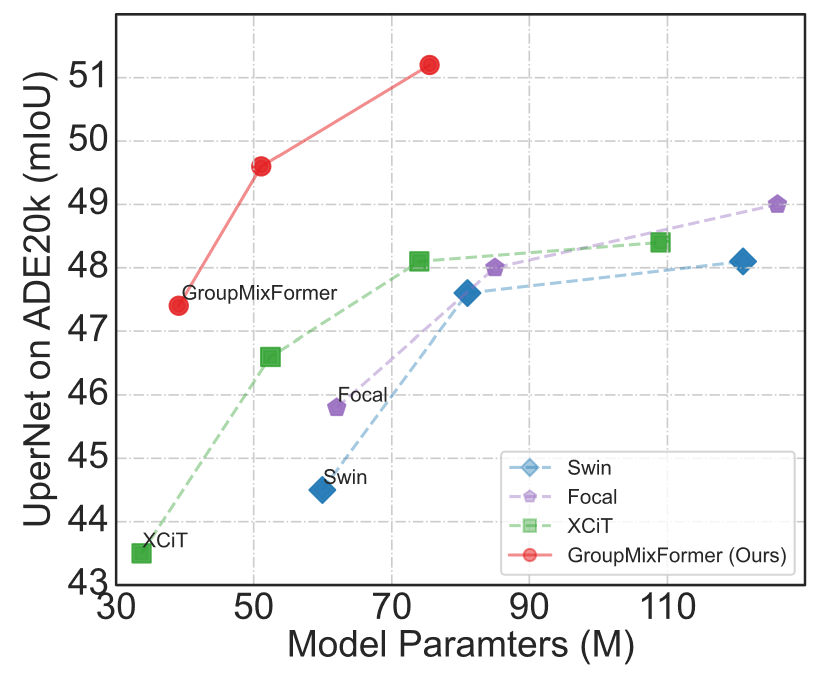} \\
				(a) Classification & (b) Detection (Mask R-CNN) & (c) Segmentation (UperNet)\\
			\end{tabular}
		\end{center}
		\vspace{-1.5em}
\caption{\textbf{Performance of \model\ compared to the state-of-the-art models.} We evaluate \model\ on standard benchmarks, including classification on ImageNet-1K~\citep{russakovsky-15ijcv-imagenet} without extra data in (a), object detection on COCO~\citep{lin-14eccv-coco} in (b), and semantic segmentation on ADE20K~\citep{zhou-19ijcv-ade20k} in (c). The computational complexity is denoted as the geometry area. \model\ performs favorably against ViT and CNN models including DeiT~\citep{touvron-21ICML-deit}, Swin~\citep{liu-iccv21-swin}, PVT~\citep{wang-iccv21-pvt}, CoaT~\citep{xu-21iccv-coat}, Focal~\citep{yang-21-focal}, ConvNeXt~\citep{liu-22-convnet}, etc.}
\label{fig:teaser_performance}
\vspace{-1.5em}
\end{figure*}

%% file: src/secs/2-related.tex
\section{Related Works}
\subsection{Vision Transformer}

Vision Transformer (ViT)~\citep{dosovitskiy-iclr21-vit} first introduces the Transformers into computer vision. Unlike CNN-based architectures, ViT utilizes sequentially-connected Transformer encoders~\citep{vaswani-nips17-trans} on the visual token sequence. The multi-head self-attention (MHSA) mechanism employed in ViTs captures global dependencies effectively, giving them an edge over CNN neural networks~\citep{he-16cvpr-resnet,huang-17cvpr-densenet} in both supervised~\citep{graham-21iccv-levit,touvron-21ICML-deit} and self-supervised scenarios~\citep{chen-21iccv-mocov3,caron-21iccv-dino}. To advance the general performance of ViTs, a series of researches have been conducted, including data-efficient training~\citep{touvron-21ICML-deit}, token re-designing and selection~\citep{rao-21nips-dynamicvit,liang-22iclr-evit}, pyramid structures~\citep{liu-iccv21-swin,wang-iccv21-pvt}, modulation on self-attention mechanism~\citep{zhang-21iccv-visionlongformer,chen-21-butterfly,chen-21-scatterbrain}, etc. Most of these works adopt the original Q-K-V computation, which is found to be effective in processing visual information. In this work, we aim to further advance the general performance of ViTs by introducing Group-Mix Attention (GMA). Unlike prior arts, \block\ is capable of modeling the correlations among not only individual tokens but also groups of tokens \textit{within each single Transformer encoder layer}, thus leading to comprehensive representational capabilities.

\subsection{Comprehensive Modeling of Self-Attention}

To enhance the representational abilities of self-attention, several approaches have been explored from different perspectives as shown in the following. (1) Introducing locality has been demonstrated effective, as exemplified by Swin Transformers.~\citep{liu-iccv21-swin,liu-2021-swinv2} and Focal Transformer~\citep{yang-21-focal}, which conduct attention computation within local windows. (2) Computing correlations with pre-defined patterns can enhance the capability of self-attention, as demonstrated by the CSWin Transformer~\citep{dong-2021-cswin} and Pixelfly-Mixer~\citep{chen2021pixelated}, both of which attempt to compute attention with pre-defined and carefully-designed patterns to realize more comprehensive modeling. (3) Other network architectures~\citep{xu-21iccv-coat,lee-21-mpvit,wang-21cvm-pvtv2,li-22-uniformer,wu2022p2t,si2022inception,wang2108crossformer,ren2022shunted} have also been investigated for the modeling of more comprehensive visual patterns.
In this work, we focus on the limitations caused by token-to-token correlations at one single granularity and propose an advanced attention mechanism (i.e., \textbf{GMA}) that constructs a more comprehensive prototype of self-attention, which clearly distinguishes our method from previous approaches.

%% file: src/secs/3-methods.tex
\section{GroupMix Attention and \model}
	
We introduce the motivation behind the high-level idea in ~\cref{sec:3.1}, elaborate on the structural designs in ~\cref{sec:3.2}, and describe the architectural configurations in ~\cref{sec:3.3}.

\subsection{Motivation: from Individual to Groups} \label{sec:3.1}

We discuss the limitations of self-attention starting from its vanilla formulation. Let $\mathrm{X}\in\mathbb{R}^{N\times d}$ be the input tokens, where $\mathit{N}$ is the token number and $\mathit{d}$ is the dimension. The output of vanilla self-attention is,
	\begin{equation}
	\mathrm{Y} = \text{Softmax}(\mathrm{X}\mathrm{X}^\text{T})\mathrm{X} \,.
	\end{equation}
Note that we ignore the normalization factor $\frac{1}{\sqrt{d}}$ for brevity. Intuitively, by the definition of matrix multiplication, $\mathrm{X}\mathrm{X}^\text{T}$ calculates the similarity/correlation between each two of the tokens. The output of the softmax function $\mathrm{A}\in\mathbb{R}^{N\times N}$ is called an attention map. The multiplication $\mathrm{A}\mathrm{X}$ means linearly re-combining the tokens according to the attention map at each location.
	
We note a limitation of this form. There may exist certain patterns (i.e., group patterns) that require treating some specific tokens as a group with diverse granularities. However, self-attention lacks an explicit mechanism for modeling such patterns, as it only considers correlations between pairs of individual tokens at a single granularity (i.e., individual patterns). In this paper, we seek to utilize both individual patterns and group patterns for comprehensive modeling. 
Unlike prior approaches that model distinct patterns across multiple stages (typically four stages in a Transformer backbone), our approach introduces a novel method of encoding this modeling process within each individual layer at each stage. Specifically, for group patterns, we seek to correlate some neighborhoods of tokens to the other neighborhoods. This paper proposes to achieve this by generating \emph{group proxies} in Query, Key, and Value, and performing the Q-K-V computation with proxies, which is described in ~\cref{sec:3.2}. We experimentally found that explicitly modeling the correlations among groups with diverse sizes and individual tokens significantly improves the performance of not only the proposed \model\ but also other ViTs with different attention modules (e.g., Swin Transformer~\citep{liu-iccv21-swin} and PVT~\citep{wang-iccv21-pvt}, as shown in ~\cref{tab:abla_others}), demonstrating that upgrading the fundamental component can benefit multiple ViTs.

\subsection{\block: Mixing Groups for Better Attention}\label{sec:3.2}

\input{src/figure_latex/ama-pipeline}

We introduce \block\ to model the group patterns as aforementioned.
In \block, we generate the group proxies by replacing some entries in the Query, Key, and Value with aggregations of some whole groups, which can be efficiently implemented with sliding-window-based operations $\operatorname{Agg(\cdot)}$, e.g., max-pooling, convolution and etc. 
Specifically, the Q/K/V entries are uniformly divided into $\mathit{n}$ segments and we perform aggregation on some segments. Without loss of generality, we use $\mathrm{X}_i$ ($i \in [1,\cdots, n]$) to denote one segment ($\mathrm{X}$ may represent $\mathrm{Q}$, $\mathrm{K}$, or $\mathrm{V}$) and the aggregations as $\operatorname{Agg}^i(\mathrm{X}_i)$. Note that the aggregator may be different for each segment. To perform attention computation, we concatenate the aggregations $\operatorname{Agg}^i(\mathrm{X}_i), i\in [1,\cdots, n]$ to produce $X^\prime$. In this way, we obtain group proxies $Q^\prime$, $K^\prime$, and $V^\prime$.
Afterward, we perform attention computation as introduced in ~\citep{xu-21iccv-coat,ali-21nips-xcit,shen-21IWCACV-efficient} on the group proxies to generate the output.

During the aggregation process, we maintain the feature resolution. Therefore, without reducing the spatial resolution, \block\ brings fine-grained features for attention computation, which outperforms those with decreased feature sizes~\citep{fan-21iccv-multiscale,wu2021cvt}. 
In this paper, we use depth-wise convolutions with various kernel sizes to implement aggregators $\operatorname{Agg}(\cdot)$, though we find other implementations also work (as shown in \cref{tab:abla_implementation}). 
As the inputs of attention are now group proxies, we achieve correlating K$\times$K tokens simultaneously (K denotes the kernel size of $\operatorname{Agg}(\cdot)$, which may be different for each segment) instead of individual tokens, which is more sufficient and comprehensive for modeling correlations. 

The idea of using sliding-window-based operations to aggregate groups into proxies, though simple, is the key to the mechanism of mixing \emph{groups of different sizes} and individual tokens at \emph{various granularities}, as we use a different kernel size of aggregator for each segment. Such a process can be efficiently implemented via splitting segments, feeding them through aggregators implemented with different kernel sizes, and concatenating the outputs. Moreover, inspired by~\citep{dong-2021-cswin}, we also employ an identity mapping on one segment instead of an aggregator to maintain the network's abilities in modeling individual token correlations. 
Therefore, we can model correlations among both groups and tokens while computing the attention map. Multiplying the attention map with the Value can be viewed as re-combining the corresponding groups together with individual tokens accordingly.

Specifically, following the implementation of self-attention~\citep{shen-21IWCACV-efficient,ali-21nips-xcit,xu-21iccv-coat}, we also use three learnable linear projections to generate Q, K, and V. Afterward, we split Q/K/V uniformly into five segments, each of which participates in different computations. As shown in ~\cref{fig:pipeline} (the left part), a branch corresponds to an aforementioned segment, and the four branches whose outputs are fed into the attention computation are referred to as the pre-attention branches. 
In three of the pre-attention branches, we use various implementations (e.g., min-pooling, avg-pooling, max-pooling, depth-wise convolution) as the aggregator $\operatorname{Agg(\cdot)}$ with different kernel sizes, which are set as 3,5,7, respectively. The results in ~\cref{tab:abla_implementation} indicate that each of these implementations achieves favorable performance, which shows that aggregation is a crucial step for attention advancement while its implementation can be flexible. We adopt the depth-wise convolutions, whose results are slightly better, in our paper.
We further diversify the structures by using no aggregator in the last pre-attention branch, making it an identity mapping. Apart from such a branch with attention but no aggregator, we construct another branch with an aggregator but no attention, which is referred to as the non-attention branch. Finally, the outputs are mixed by a token ensemble layer, which is simply implemented by a linear projection with normalization~\citep{ba-16-layer} and activation.

\input{src/tables/architecture}

\subsection{Architectural Configurations}\label{sec:3.3}
Building on the proposed Group-Mix Attention, we introduce a series of vision Transformers named \model, as shown in ~\cref{fig:pipeline}. We adopt a hierarchical~\citep{liu-iccv21-swin,wang-iccv21-pvt} topology with four stages. The first 4$\times$ patch embedding layer embeds images into tokens, which is implemented with two sequential 3$\times$3 convolutional layers, each with a stride of 2 and another two 3$\times$3 layers with a stride of 1. At the beginning of each last three stages, we use a 2$\times$ patch embedding, which is also implemented with a 3$\times$3 convolution. Within each stage, we construct several encoder blocks. Apart from a \block\ block introduced in the last subsection, an encoder block also contains a Feed-Forward Network (FFN), Layer Normalization~\citep{ba-16-layer} and identity shortcuts, 
following the common practice in~\citep{dosovitskiy-iclr21-vit,touvron-21ICML-deit,liu-iccv21-swin,wang-iccv21-pvt,yang-21-focal}. For image classification, the final output tokens are fed into the classifier after global average pooling (GAP); for dense prediction tasks (e.g., object detection and semantic segmentation), the task-specific heads can utilize the pyramid features output by the four stages. We do not adopt positional encoding in our model since we have naturally broken the permutation invariance with the \block\ aggregators.
	
We instantiate four models with different architectural configurations. The architectural hyper-parameters include the number of encoder blocks in each stage $\mathit{L}$, the embedded dimension $\mathit{D}$, and the MLP ratio $\mathit{R}$, as shown in ~\cref{tab:architecture}. Following the prior works~\citep{wang-iccv21-pvt,liu-iccv21-swin,touvron-21ICML-deit}, our models scale up from the mobile-scale \model-M (5.7 M) to the large-scale \model-L (70.3 M).

%% file: src/figure_latex/ama-pipeline.tex
\begin{figure*}[t]
    \centering
    \begin{tabular}{c}
        \includegraphics[width=0.88\linewidth]{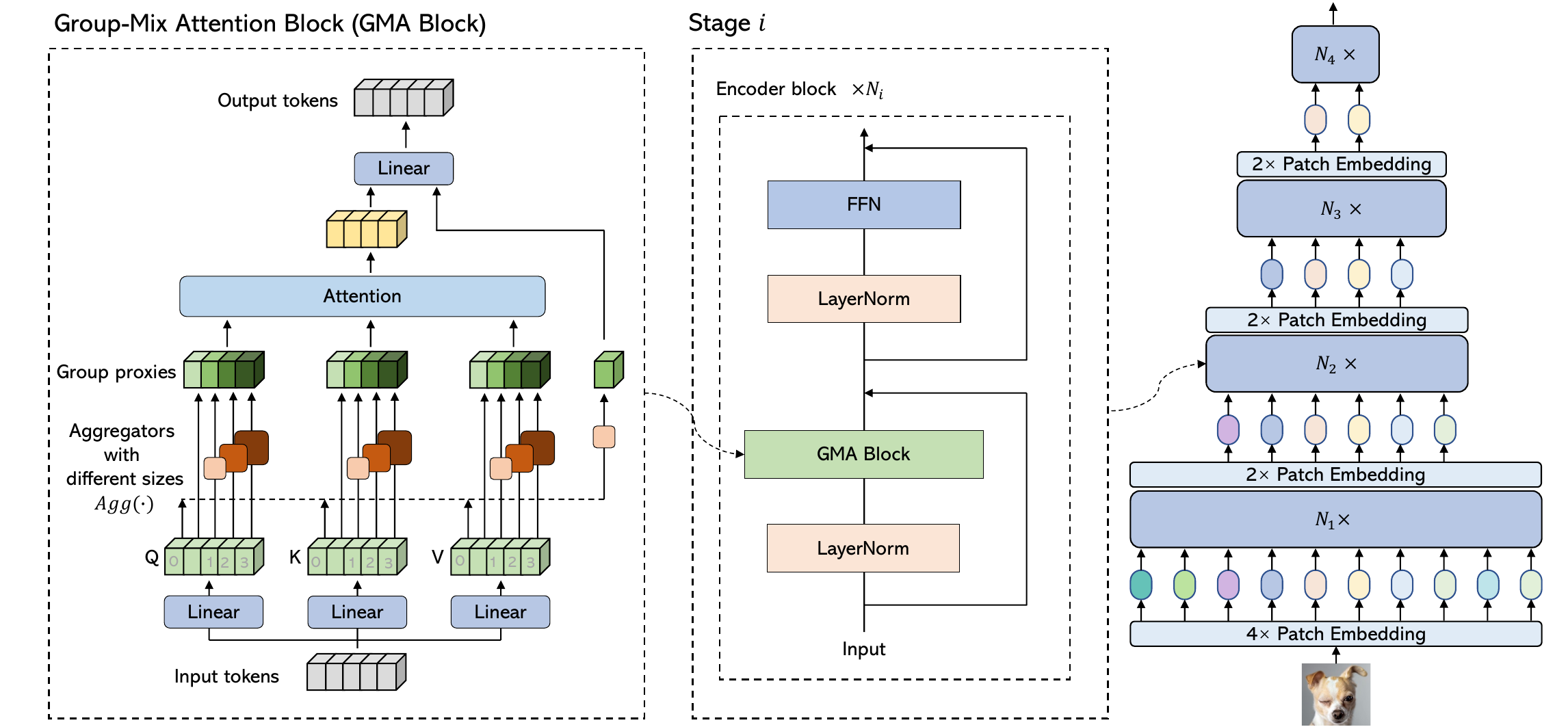}            
    \end{tabular}
\vspace{-1.em}
    \caption{\textbf{Structural designs of Group-Mix Attention Block and architecture of \model.} In each \block\ block, we split $\mathit{Q}$, $\mathit{K}$, and $\mathit{V}$ into five segments and use aggregators with different kernel sizes to generate group proxies on four of them, so that we can conduct attention computation on mixtures of individual tokens and group proxies of different granularities. The branches whose outputs are fed into the attention computation are referred to as \textit{\textbf{the pre-attention branches}}. To construct diverse connections, the rightmost branch utilizes aggregation but without attention, which is termed \textit{\textbf{the non-attention branch}}. A linear mapping layer is adopted to fuse the outputs from the attention and non-attention branch. For clear illustration, we use $\operatorname{Agg}^1$, $\operatorname{Agg}^2$, and $\operatorname{Agg}^3$ in the pre-attention branch to denote the aggregators with kernel sizes of 3, 5, and 7, respectively, and use $\operatorname{Agg}^0$ for the aggregator in the non-attention branch. }
    \label{fig:pipeline}
\vspace{-1.5em}
\end{figure*}

%% file: src/tables/architecture.tex
\begin{table*}[t]
		\centering
		\renewcommand{\tabcolsep}{4.4mm}
  		\scriptsize
		\caption{
			Architectural configurations of \model\ models. We use D, R, and L to denote the dimension of tokens, the expansion ratio of FFN, and the number of encoder blocks. We use M/T/S/B/L (mobile/tiny/small/base/large) to label models with different scales.
		}

		\label{tab:architecture}
		\begin{tabular}{c|c|c|c|c|c|c}
			\hline
			
			& Output size &  \makecell{\model\\-M(5.7M)} &  \makecell{\model\\-T(10.9M)} & \makecell{\model\\-S(22.4M)} & \makecell{\model\\-B(45.8M)} & \makecell{\model\\-L(70.3M)} \\
			\hline
			
			\makecell{stage 1} & 
			$\frac{H}{4}\times \frac{W}{4}\times D_1$ & 
			\makecell{$\mathit{D}_1=40$\\$\mathit{R}_1=4,\mathit{L}_1=3$} & 
			\makecell{$\mathit{D}_1=80$\\$\mathit{R}_1=4,\mathit{L}_1=4$} & 
			\makecell{$\mathit{D}_1=80$\\$\mathit{R}_1=4,\mathit{L}_1=2$} & 
			\makecell{$\mathit{D}_1=200$\\$\mathit{R}_1=2,\mathit{L}_1=8$} & 
			\makecell{$\mathit{D}_1=240$\\$\mathit{R}_1=4,\mathit{L}_1=8$} \\
			\hline
			\makecell{stage 2} & 
			$\frac{H}{8}\times \frac{W}{8}\times D_2$ & 
			\makecell{$\mathit{D}_2=80$\\$\mathit{R}_2=4, \mathit{L}_2=3$} & 
			\makecell{$\mathit{D}_2=160$\\$\mathit{R}_2=4, \mathit{L}_2=4$} & 
			\makecell{$\mathit{D}_2=160$\\$\mathit{R}_2=4, \mathit{L}_2=4$} & 
			\makecell{$\mathit{D}_2=240$\\$\mathit{R}_2=2, \mathit{L}_2=8$} & 
			\makecell{$\mathit{D}_2=320$\\$\mathit{R}_2=4, \mathit{L}_2=10$} \\
			
			\hline
			\makecell{stage 3} & 
			$\frac{H}{16}\times \frac{W}{16}\times D_3$ & 
			\makecell{$\mathit{D}_3=160$\\$\mathit{R}_3=4, \mathit{L}_3=12$} & 
			\makecell{$\mathit{D}_3=200$\\$\mathit{R}_3=4, \mathit{L}_3=12$} & 
			\makecell{$\mathit{D}_3=320$\\$\mathit{R}_3=4, \mathit{L}_3=12$} & 
			\makecell{$\mathit{D}_3=320$\\$\mathit{R}_3=4, \mathit{L}_3=12$} & 
			\makecell{$\mathit{D}_3=360$\\$\mathit{R}_3=2, \mathit{L}_3=30$} \\
			
			\hline
			\makecell{stage 4} & 
			$\frac{H}{32}\times \frac{W}{32}\times D_4$ & 
			\makecell{$\mathit{D}_4=160$\\$\mathit{R}_4=4, \mathit{L}_4=4$} &
			\makecell{$\mathit{D}_4=240$\\$\mathit{R}_4=4, \mathit{L}_4=4$} &
			\makecell{$\mathit{D}_4=320$\\$\mathit{R}_4=4, \mathit{L}_4=4$} &
			\makecell{$\mathit{D}_4=480$\\$\mathit{R}_4=4, \mathit{L}_4=8$} &
			\makecell{$\mathit{D}_4=480$\\$\mathit{R}_4=2, \mathit{L}_4=10$} \\
			
			\hline
		\end{tabular}
        \vspace{-1.5em}
	\end{table*}	

%% file: src/secs/4-exp.tex
\section{Experiments}
In this section, we evaluate our \model\ on standard visual recognition benchmarks including ImageNet-1K~\citep{russakovsky-15ijcv-imagenet}, MS-COCO~\citep{lin-14eccv-coco}, and ADE20k~\citep{zhou-19ijcv-ade20k}. We present the implementation details for each scenario, quantitative comparisons to state-of-the-art vision backbones, and ablation studies in the following. 

\subsection{Implementation Details} \label{sec:implementation}
We evaluate the image classification performance of \model\ on the ImageNet-1K dataset. We follow~\citep{zhang-17-mixup,yun-19iccv-cutmix,touvron-21ICML-deit} to augment data and use the training recipe in~\citep{liu-iccv21-swin}. We train \model\ for 300 epochs using an initial learning rate of $\mathrm{10}^{-3}$ with a 20-epoch linear warm-up. AdamW optimizer~\citep{loshchilov-17-adamw} is utilized with a weight decay of 0.05 and a cosine learning rate schedule. The stochastic depth drop rates~\citep{huang-16eccv-dpr} are set to 0.0, 0.1, 0.2, 0.4, and 0.5 for \model-M/T/S/B/L, respectively. For higher resolutions (e.g., $\mathrm{384}^2$ or $\mathrm{448}^2$), we finetune the models in another 30 epochs with the learning rate initialized as $2\times\mathrm{10}^{-6}$ and a linear warm-up for 5 epochs. The finetuning process uses AdamW~\citep{loshchilov-17-adamw} with a weight decay of $\mathrm{10}^{-8}$ for optimization.

For object detection and instance segmentation, COCO 2017 dataset is utilized. Specifically, we employ \model\ as the backbones of Mask R-CNN~\citep{he-17iccv-maskrcnn} for object detection and segmentation, and RetinaNet~\citep{lin-17iccv-retinanet} for detection only. All the backbones are initialized via the corresponding ImageNet pretrained models. We follow the training schedules in~\citep{mmdetection}: the initial learning rate is set to $\mathrm{10}^{-4}$ with a linear warm-up for 500 iterations and gradually decreases to $\mathrm{10}^{-5}$ and $\mathrm{10}^{-6}$ at the 24-\textit{th} and 33-\textit{th} epochs, respectively. We use AdamW~\citep{loshchilov-17-adamw} for both Mask R-CNN and RetinaNet, but the weight decay is 0.05 for the former and $\mathrm{10}^{-4}$ for the latter. Except for COCO, we also evaluate the semantic segmentation performance on ADE20k with UperNet~\citep{xiao-18eccv-upernet} and Semantic FPN~\citep{kirillov-19cvpr-fpn}. We follow~\citep{wang-iccv21-pvt,liu-iccv21-swin} to use the public toolkit~\citep{mmseg2020} for training and evaluations. The Semantic FPN is trained for 80k iterations, while the UperNet is trained for 160k iterations, both with an AdamW optimizer.

\subsection{Comparisons with State-of-the-art Models}
{\flushleft \bf Image Classification.}
We compare the proposed \model\ with the state-of-the-art models from the literature in ~\cref{tab:imagenet_cls}, where all the reported results use only ImageNet-1k for training. For fair comparisons, we do not use any extra augmentations, like token-labeling~\citep{jiang-21nips-tokenlabeling}, knowledge distillation, SAM~\citep{foret-2020-sam}, etc.  We observe that \model\ consistently achieves higher Top-1 accuracies than the ViT and CNN models under similar model sizes and computational complexity constraints. Specifically, tested with a resolution of $\mathrm{224}^2$, \model-S yields a top-1 accuracy of 83.4\% with only 22.4M parameters, outperforming the second best ViT (CSWin-T~\citep{dong-2021-cswin}) by 0.7\% and the best CNN (ConvNext-T~\citep{liu-22-convnet}) by 1.3\%. Meanwhile, \model-B trained with $224\times244$ images even achieves a similar accuracy with Swin-B~\citep{liu-iccv21-swin}, though the size of \model-B is only half as that of Swin-B. Moreover, \model\ shows satisfying scalability towards higher resolution. For example, finetuning with a resolution of $\mathrm{384}^2$ further improves the performance of \model-S to 85.0\%; with around 70M parameters, our \model-L achieves 85.0\% with a resolution of $\mathrm{224}^2$ and 86.2\% with $\mathrm{384}^2$. 
The results underscore the advantages of comprehensively incorporating both token-to-token and group-to-group correlations in modeling visual patterns. Additionally, attention responses from different aggregators are presented in the appendix to support the notion that there exist some patterns so that some tokens should be handled as a whole for classification.

Moreover, we empirically observed that implementing depth-wise convolutions as the aggregators in GMA does lead to a slowdown in the inference speed. The throughput is reported in the appendix. However, this could be improved with more efficient aggregators (e.g., avg-pooling) and  implementing engineering optimizations, such as \textit{``torch.compile''}. We will explore the optimization of the model's real-world speed in future research.

\input{src/tables/exp-classification}

{\flushleft \bf Object Detection.} ~\cref{tab:maskrcnn-3x} shows the object detection results on COCO with Mask R-CNN and RetinaNet detectors. With Mask R-CNN, \model\ achieves higher average precision under similar model parameters. Specifically, \model-T performs 1.0\% higher (i.e., 47.5\% v.s. 46.5\%) than the second-best model, which is CoaT Mini, while maintaining a smaller model size of 30.8 M. Besides, our \model-B achieves an $\mathrm{AP}^b$ of 51.5\%, surpassing all the comparable models. With RetinaNet, \model\ also shows superiority: \model-T performs 0.5\% better than Swin-B (i.e., 46.3\% v.s. 45.8\%) though ours is much smaller (i.e., 20.2 M v.s. 98.0 M); \model-B performs 2.9\% better (i.e., 50.2\% v.s. 47.3\%) than the second best mdoel, i.e., Focal-small. These results show that \model\ achieves favorable performance with both detectors. The consistent and significant improvements demonstrate the effectiveness of the Group-Mix mechanism, which is supposed to be able to capture the fine-grained features to facilitate dense predictions.

\input{src/tables/exp-detection}

\input{src/tables/exp-segmentation}

{\flushleft \bf Semantic Segmentation.} ~\cref{tab:maskrcnn-3x} also shows the semantic segmentation results on COCO with Mask-RCNN. Our \model-T impressively achieves an $\mathrm{AP}^m$ of 42.4\%, 0.6\% higher than Coat Mini and 1.7\% higher than PVT-Large. Besides, \model-B performs 1.1\% better than Uniformer-B (i.e., 45.9\% v.s. 44.8\%). On ADE20K, we use UperNet and Semantic FPN and report the results in ~\cref{tab:ade20k}. Similarly, we observe that \model s consistently achieve favorable performance compared to the existing backbones. For example, \model-T, though much smaller, performs 2.0\% better than XCiT-S12/8 (i.e., 46.2\% v.s. 44.2\%, 14.1 M v.s. 30.4 M) with Semantic FPN. Notably, \model-T outperforms XCiT-M24/16 by 0.3\%, though the latter is 6.4$\times$ as big as \model-T (i.e., 46.2\% v.s. 45.9\%, 14.1 M v.s. 90.8 M). Similarly, with UperNet, \model s perform much better than the other bigger models, showing a clearly better trade-off between performance and efficiency. Such significant improvements suggest that the Group-Mix mechanism is able to produce high-quality features for pixel-level predictions.

\subsection{Ablation Studies} \label{sec:abla}
In this sub-section, we conduct ablation studies to analyze the key designs of \model . (1) We first analyze the necessity of the aggregators by changing the structural designs of GMA. (2) We experiment with various implementations of aggregators to see if other sliding-window-based operations, except for convolution, also work. (3) We validate that the performance gains of \model\ do not stem from the macrostructures. (4) The optimal configurations of the kernel sizes have been explored. (5) We conduct experiments to verify that GMA is not merely a trivial combination of convolution and self-attention. (6) We plug GMA Blocks into the other popular ViT architectures to verify if the superior performance of \model\ is merely due to the architectural designs (e.g., overlapping embedding layers and numbers of blocks within each stage).
For image classification, we train \model-T for 300 epochs on ImageNet-1k ($\mathrm{224}^{2}$) and test with the validation set. For object detection and semantic segmentation, we train Mask R-CNN with the 1$\times$ schedule~\citep{mmdetection} on COCO.

\input{src/tables/exp-abla-aggregators}

{\flushleft \bf Group aggregators are necessary.} \cref{tab:abla_aggmodule} shows the results of ablating the aggregators. We first construct a \model-T baseline by replacing all of the five branches in GMA Blocks with identity mappings, so that the block degrades into a regular self-attention module. In the first group of experiments, we restore the aggregators in the non-attention branch ($\operatorname{Agg}^0$) or the three pre-attention branches ($\operatorname{Agg}^{1}$, $\operatorname{Agg}^{2}$ and $\operatorname{Agg}^{3}$). Every model is trained from scratch with the same configurations as described in \cref{sec:implementation}. It could be observed that the aggregators are all critical, as they improve the top-1 accuracy by 0.4\% and 1.0\%, respectively. 

Moreover, the second group of experiments in \cref{tab:abla_aggmodule} shows that using aggregators in all of the three pre-attention branches yields better performance than using any single one. Similar experimental results are observed in object detection and semantic segmentation as well. Using all the aggregators improves the baseline performance by a certain margin (e.g., +0.7\% $\mathrm{AP}^b$ and +0.5\% $\mathrm{AP}^m$). These results indicate that modeling correlations in a more comprehensive manner is able to provide fine-grained visual representations to benefit dense prediction scenarios. 

\input{src/tables/exp-abla-aggregator-implementation}

\input{src/tables/exp-abla-gma-necessary}

\input{src/tables/exp-abla-kernel-configuration}

\input{src/tables/exp-abla-gma-generalization}

We then analyze the impact of various kernel sizes of pre-attention aggregators on performance. Without altering the non-attention branch, we replace all of the pre-attention aggregators with either $\operatorname{Agg}^1$ (3$\times$3 convolution), $\operatorname{Agg}^2$ (5$\times$5) or $\operatorname{Agg}^3$ (7$\times$7). 
The second set of results in \cref{tab:abla_aggmodule} indicates that the utilization of any group aggregators enhances classification and dense prediction performance, with a diverse combination of 3$\times$3, 5$\times$5, and 7$\times$7 yielding the most optimal results. Specifically, 
\model-T equipped with diverse aggregators outperforms the baseline by +1.6\% classification accuracy, +1.5\% $\mathrm{AP}^b$ in object detection, 
and +1.0\% $\mathrm{AP}^m$ in semantic segmentation, which suggests that modeling the correlations among groups of \emph{diverse} sizes is the key to performance boost.

{\flushleft \bf Depthwise Convolutions are effective aggregators.} Note that the implementations of aggregators $\operatorname{Agg}(\cdot)$ could be various. \cref{tab:abla_implementation} shows our results regarding the effects of different aggregator implementations (e.g., depthwise convolution~\citep{chollet-17cvpr-dwconv}, max-pooling, or average-pooling). It's empirically observed that the aggregators implemented by depthwise convolution achieve the slightly better performance (82.5\% Top-1 accuracy on classification, 42.5\% $\mathrm{AP}^b$ for detection, and 39.7\% $\mathrm{AP}^m$ for instance segmentation with Mask R-CNN). Compared with the max-pooling and min-pooling operations, convolutional aggregators may take advantage of involving more learnable parameters for computing correlations, thus achieving better performances. 

{\flushleft \bf Performance gains are not derived from macro-structures.} Compared with the representative works~\citep{liu-iccv21-swin,wang-iccv21-pvt,touvron-21ICML-deit}, our \model\ is deeper and has different implementations of patch embedding. In order to justify that the performance gains are not simply due to a better combination of architectural hyper-parameters (including the dimensions of tokens, expansion ratios, and layer depths as introduced in ~\cref{tab:architecture}), we replace the GMA Blocks in \model-T with the Swin-attention or PVT-attention. The results in ~\cref{tab:abla_marco} show that simply replacing the GMA causes a significant performance drop, which justifies that the performance gain is due to the advanced attention mechanism instead of the architecture.

{\flushleft \bf Optimal configurations on the kernel sizes of aggregators.} To find the optimal configuration, we undertake two approaches: (1) enlarging the kernel size, and (2) altering the kernel configurations in varying orders. The first approach entails increasing the kernel sizes from (3,5,7) to (5,7,9). For the second approach, we deploy aggregators with larger kernels in the shallow layers and smaller kernels in the deeper layers, as well as in a reversed configuration. However, as demonstrated in ~\cref{tab:abla_kernel_configuration}, neither of these modifications proved as effective as the configuration we ultimately adopted.

{\flushleft \bf GMA is not merely a trivial combination of convolution and self-attention.} We conduct further experiments to validate that our proposed \model\ is essentially different from a simple combination of convolution and self-attention. Specifically, we remove all the group aggregators from \model-T and insert a group of convolutional layers organized in the same manner (i.e., a combination of parallel identity mapping, 3$\times$3, 5$\times$5 and 7$\times$7 layers) before the whole self-attention module. The accuracy drops by 1.0\% in the Top-1 accuracy (81.5\% v.s. 82.5\%). 

{\flushleft \bf Aggregator is an advanced universal building block that could be applied to the other ViTs.} We may also incorporate aggregators into representative ViTs (e.g., Swin~\citep{liu-iccv21-swin} and PVT~\citep{wang-iccv21-pvt}) by simply inserting the them into their original attention modules to process their Query, Key, and Value. 
The results in \cref{tab:abla_others} show that such a strategy generally boosts ViTs by a clear margin. For example, PVT-Small with aggregators achieves 80.6\% Top-1 accuracy, which is 0.8\% higher than its original result. It indicates that the proposed aggregators advance ViTs by modeling the group correlations and thus leading to a comprehensive understanding of the tokens.

%% file: src/tables/exp-classification.tex
\begin{table}[t!]
\centering
\renewcommand{\arraystretch}{1}
\renewcommand{\tabcolsep}{1.2mm}
\scriptsize
\caption{\textbf{ImageNet-1k validation accuracy.} The GFLOPs are measured with the specific  resolution. Models with a comparable number of parameters are grouped together.}
\vspace{-0.5em}
\begin{tabular}{l|c|c|cc|c}
    \hline
	Method & Type & \#Params.(M) & Input & \#GFLOPs & Top-1 (\%)  \\
	\hline
	ShuffleNet v2-50~\citep{ma-eccv18-shufflenetv2} & CNN & 2.3 & $\mathrm{224}^2$ & 2.3 & 77.2 \\
	Mobile-Former~\citep{chen-22cvpr-mobileformer} & Trans & 4.6 & $\mathrm{224}^2$ & 1.2 & 72.8 \\
	MobileViT-S~\citep{mehta-2021-mobilevit} & Trans & 5.6 & $\mathrm{256}^2$ & 1.8 & 78.4 \\
	\sota{\model-M} & \sota{Trans} & \sota{5.7} & \sota{$\mathrm{224}^2$} & \sota{1.4} &\sota{\textbf{79.4}} \\
	\sota{\model-M} & \sota{Trans} & \sota{5.7} & \sota{$\mathrm{384}^2$} & \sota{4.0} &\sota{\textbf{81.5}} \\
	\sota{\model-M} & \sota{Trans} & \sota{5.7} & \sota{$\mathrm{448}^2$} & \sota{5.4} &\sota{\textbf{81.8}} \\
	\hline
	ResNet18~\citep{he-16cvpr-resnet} & CNN & 11.7 & $\mathrm{224}^2$ &1.8 & 69.8  \\
     EffNet-B4~\citep{freeman2018effnet} & CNN & 19.0 & $\mathrm{224}^2$ & 4.2 & 82.9\\
	PVT-Tiny~\citep{wang-iccv21-pvt} & Trans & 13.2 & $\mathrm{224}^2$ & 1.9 & 75.1 \\
	PVTv2-B1~\citep{wang-21cvm-pvtv2} & Trans & 13.1 & $\mathrm{224}^2$ & 2.1 & 78.7 \\
    P2T-Tiny~\citep{wu2022p2t} & Trans & 11.6 & $\mathrm{224}^2$ & 1.8 & 79.8 \\
    CoaT Mini~\citep{xu-21iccv-coat} & Trans & 10.0 & $\mathrm{224}^2$ & 6.8 & 81.0 \\
    BiFormer-T~\citep{zhu2023biformer} & Trans & 13.1 & $\mathrm{224}^2$ & 2.2 & 81.4 \\

	\sota{\model-T} & \sota{Trans} & \sota{11.0} & \sota{$\mathrm{224}^2$} & \sota{3.7} &
	\sota{\textbf{82.5}} \\
	\sota{\model-T} & \sota{Trans} & \sota{11.0} & \sota{$\mathrm{384}^2$} & \sota{10.9} & \sota{\textbf{84.1}} \\
	\sota{\model-T} & \sota{Trans} & \sota{11.0} & \sota{$\mathrm{448}^2$} & \sota{14.9} & \sota{\textbf{84.3}} \\

	\hline
	ResNet50~\citep{he-16cvpr-resnet} & CNN  & 25.6 & $\mathrm{224}^2$ & 4.1 & 76.5 \\
	ResNeXt50-32x4d~\citep{xie-17cvpr-resnext} & CNN & 25.0 & $\mathrm{224}^2$ & 4.3 & 77.6 \\
	ConvNeXt-T~\citep{liu-22-convnet} & CNN & 29.0 & $\mathrm{224}^2$ & 4.5 & 82.1 \\
	PVT-Small~\citep{wang-iccv21-pvt} & Trans & 24.5 & $\mathrm{224}^2$ & 3.8 & 79.8 \\
	PVTv2-B2~\citep{wang-21cvm-pvtv2} & Trans & 25.4 & $\mathrm{224}^2$ & 4.0 & 82.0 \\
	Swin-T~\citep{liu-iccv21-swin} & Trans & 29.0 & $\mathrm{224}^2$ & 4.5 & 81.3 \\
	CoaT Small~\citep{xu-21iccv-coat} & Trans & 22.0 & $224^2$ & 12.6 & 82.1 \\
	Focal-Tiny~\citep{yang-21-focal} & Trans & 29.1 & $224^2$ & 4.9 & 82.2 \\
    P2T-Small~\citep{wu2022p2t} & Trans & 24.1 & $\mathrm{224}^2$ & 3.7 & 82.4 \\
    CSWin-T~\citep{dong-2021-cswin} & Trans &  23.0 & $\mathrm{224}^2$ & 4.3  & 82.7 \\
    MViTv2-T~\citep{li2022mvitv2} & Trans & 24.0 & $\mathrm{224}^2$ & 4.7 & 82.3 \\
    DaViT-T~\citep{ding2022davit} & Trans  & 28.3 & $\mathrm{224}^2$ & 4.5 & 82.8 \\
    XCiT-S12/16~\citep{ali-21nips-xcit} & Trans & 26.0 & $224^2$ & 4.8 & 82.0 \\
	\sota{\model-S} & \sota{Trans} & \sota{22.4} & \sota{$224^2$} & \sota{5.2} &\sota{\textbf{83.4}} \\
	\sota{\model-S} & \sota{Trans} & \sota{22.4} & \sota{$\mathrm{384}^2$} & \sota{15.2} &
	\sota{\textbf{85.0}}  \\
 
	\hline
	ResNet101~\citep{he-16cvpr-resnet} & CNN & 44.7 & $\mathrm{224}^2$ & 7.9 & 77.4 \\
	ResNeXt101-32x4d~\citep{xie-17cvpr-resnext} & CNN & 44.2 & $\mathrm{224}^2$ & 8.0 & 78.8 \\
	ConvNeXt-S~\citep{liu-22-convnet} & CNN & 50.0 & $\mathrm{224}^2$ & 8.7 & 83.1 \\
	ConvNeXt-B~\citep{liu-22-convnet} & CNN & 89.0 & $\mathrm{224}^2$ & 15.4 & 83.8 \\
	ConvNeXt-L~\citep{liu-22-convnet} & CNN & 198.0 & $\mathrm{224}^2$ & 34.4 & 84.3 \\
	PVT-Large~\citep{wang-iccv21-pvt} & Trans & 61.4 & $\mathrm{224}^2$ & 9.8 & 81.7 \\
	PVTv2-B3~\citep{wang-21cvm-pvtv2} & Trans & 45.2 & $\mathrm{224}^2$ & 6.9 & 83.2 \\
	Swin-B~\citep{liu-iccv21-swin} & Trans & 88.0 & $\mathrm{224}^2$ & 15.4 & 83.5 \\
	Swin-B~\citep{liu-iccv21-swin} & Trans & 88.0 & $\mathrm{384}^2$ & 47.0 & 84.5 \\
    CSWin-B~\citep{dong-2021-cswin} & Trans &  78.0 & $\mathrm{224}^2$ & 15.0 & 84.2 \\
    MViTv2-B~\citep{li2022mvitv2} & Trans  & 78.0 & $\mathrm{224}^2$ & 15.0 & 84.2 \\
    DaViT-B~\citep{ding2022davit} & Trans  & 87.9 & $\mathrm{224}^2$ & 15.5 & 84.6 \\
    MaxViT-S~\citep{tu2022maxvit} & Trans & 69.0  & $\mathrm{224}^2$ & 11.7 & 84.5 \\
	CoaTLite Medium~\citep{xu-21iccv-coat} & Trans & 45.0 & \textbf{$\mathrm{384}^2$} & 28.7 & 84.5 \\
	Focal-Small~\citep{yang-21-focal} & Trans & 51.1 & $\mathrm{224}^2$ & 9.1 & 83.5 \\
	Focal-Base~\citep{yang-21-focal} & Trans & 89.8 & $\mathrm{224}^2$ & 16.0 & 83.8 \\
     P2T-Large~\citep{wu2022p2t} & Trans & 54.5 & $\mathrm{224}^2$ & 9.8 & 83.9 \\
	XCiT-M24/8~\citep{ali-21nips-xcit} & Trans & 84.0 & $\mathrm{224}^2$ & 63.9 &  83.7 \\

	\sota{\model-B} & \sota{Trans} & \sota{45.8} & \sota{$\mathrm{224}^2$} &  \sota{17.6} & \sota{\textbf{84.7}} \\
	\sota{\model-B} & \sota{Trans} & \sota{45.8} & \sota{$\mathrm{384}^2$} & \sota{51.6} &
	\sota{\textbf{85.8}} \\

	\sota{\model-L} & \sota{Trans} & \sota{70.3} & \sota{$\mathrm{224}^2$} &  \sota{36.1} & \sota{\textbf{85.0}} \\
	\sota{\model-L} & \sota{Trans} & \sota{70.3} & \sota{$\mathrm{384}^2$} &  \sota{106.2} & \sota{\textbf{86.2}} \\

\hline
\end{tabular}
\label{tab:imagenet_cls}
\vspace{-2em}
\end{table}

%% file: src/tables/exp-detection.tex
\begin{table*}[!t]
\centering
\scriptsize
\caption{\textbf{Object detection and instance segmentation on COCO 2017~\citep{lin-14eccv-coco} with Mask R-CNN~\citep{he-17iccv-maskrcnn} and RetinaNet~\citep{lin-17iccv-retinanet}.} All the models are pre-trained on ImageNet-1K~\citep{russakovsky-15ijcv-imagenet}. `P' represents the parameter number, and `MS' denotes multi-scale training. The 3x schedule strictly follows~\citep{mmdetection}. }
\vspace{-1em}
\resizebox{0.99\linewidth}{!}{
\begin{tabular}{l|c|cccccc|cccccc}
\hline
\renewcommand{\arraystretch}{0.1}
\multirow{2}{*}{Backbone} &\multicolumn{7}{c|}{Mask R-CNN 3$\times$ + MS} & \multicolumn{6}{c}{RetinaNet 3$\times$ + MS} \\
\cline{2-14} 
 & \#P (M) & $\mathrm{AP}^b$ & $\mathrm{AP}^b_{50}$ & $\mathrm{AP}^b_{75}$ & $\mathrm{AP}^m$ & $\mathrm{AP}^m_{50}$ & $\mathrm{AP}^m_{75}$ 
 & $\mathrm{AP}^b$ & $\mathrm{AP}^b_{50}$ & $\mathrm{AP}^b_{75}$ & $\mathrm{AP}_S^b$ & $\mathrm{AP}_M^b$ & $\mathrm{AP}_L^b$ \\
\hline
ResNet18~\citep{he-16cvpr-resnet} & 31.2/21.3 & 36.9 & 57.1 & 40.0 & 33.6 & 53.9 & 35.7 & 35.4 & 53.9 & 37.6 & 19.5 & 38.2 & 46.8\\
PVT-Tiny~\citep{wang-iccv21-pvt} & 32.9/23.0 & 39.8 & 62.2 & 43.0 & 37.4 & 59.3 & 39.9 & 39.4 & 59.8 & 42.0 & 25.5 & 42.0 & 52.1\\
CoaT Mini~\citep{xu-21iccv-coat} & 30.0/-- & 46.5 & 67.9 & 50.7 & 41.8 & 65.3 & 44.8 & -- & -- & -- & -- & -- & -- \\
CoaT-Lite Mini~\citep{xu-21iccv-coat} & 31.0/-- & 42.9 & 64.7 & 46.7 & 38.9 & 61.6 & 41.7 & -- & -- & -- & -- & -- & -- \\
\sota{\model-T} & \sota{30.8/20.2} & \sota{\textbf{47.5}} & \sota{68.9} & \sota{52.2} & \sota{\textbf{42.4}} & \sota{66.1} & \sota{45.9} & \sota{\textbf{46.3}} & \sota{67.6} & \sota{49.4} & \sota{32.0} & \sota{50.3} & \sota{59.9}\\
\hline
ResNet50~\citep{he-16cvpr-resnet} & 44.2/37.7 & 41.0 & 61.7 & 44.9 & 37.1 & 58.4 & 40.1 & 39.0 & 58.4 & 41.8 & 22.4 & 42.8 & 51.6 \\
PVT-Small~\citep{wang-iccv21-pvt} & 44.1/34.2 & 43.0 & 65.3 &  46.9 & 39.9 & 62.5 & 42.8 & 42.2 & 62.7 &  45.0 & 26.2 & 45.2 & 57.2\\
Swin-T~\citep{liu-iccv21-swin} & 48.0/39.0 & 46.0 & 68.1 & 50.3 & 41.6 & 65.1 & 44.9 & 45.0 & 65.9 & 48.4 & 29.7 & 48.9 & 58.1\\
Focal-Tiny~\citep{yang-21-focal} & 48.8/39.4 & 47.2 & 69.4 & 51.9 & 42.7 & 66.5 & 45.9 & 45.5 & 66.3 & 48.8 & 31.2 & 49.2 & 58.7\\
CoaT S~\citep{xu-21iccv-coat} & 42.0/-- & 49.0 & 70.2 & 53.8 & 43.7 & 67.5 & 47.1 & -- & -- & -- & -- & -- & --\\
CoaT-Lite S~\citep{xu-21iccv-coat} & 40.0/-- & 45.7 & 67.1 & 49.8 & 41.1 & 64.1 & 44.0 & -- & -- & -- & -- & -- & --\\
Unoformer-S$_{h14}$~\citep{li-22-uniformer} & 41.0/-- & 48.2 & 70.4 & 52.5 & 43.4 & 67.1 & 47.0 & -- & -- & -- & -- & -- & --\\ 
\sota{\model-S} & \sota{42.2/31.9} & \sota{\textbf{49.1}} & \sota{70.2} & \sota{53.7} & \sota{\textbf{43.5}} & \sota{67.4} & \sota{47.3} 
& \sota{\textbf{47.6}} & \sota{68.5} & \sota{51.3} & \sota{33.1} & \sota{51.2} & \sota{61.3}\\
\hline
ResNet101~\citep{he-16cvpr-resnet} & 63.2/56.7  & 42.8 & 63.2 & 47.1 & 38.5 & 60.1 & 41.3 & 40.9 & 60.1 & 44.0 & 23.7 & 45.0 & 53.8 \\
ResNeXt101-32x4d~\citep{xie-17cvpr-resnext} & 62.8/56.4 & 44.0 & 64.4 & 48.0 & 39.2 & 61.4 & 41.9 & 41.4 & 61.0 & 44.3 & 23.9 & 45.5 & 53.7 \\
ResNeXt101-64x4d~\citep{xie-17cvpr-resnext} & 101.9/95.5 & 44.4 & 64.9 & 48.8 & 39.7 & 61.9 & 42.6 & 41.8 & 61.5 & 44.4 & 25.2 & 45.4 & 54.6\\
PVT-Medium~\citep{wang-iccv21-pvt} & 63.9/53.9 & 44.2 & 66.0 & 48.2 & 40.5 & 63.1 & 43.5 & 43.2 & 63.8 & 46.1 & 27.3 & 46.3 & 58.9 \\
PVT-Large~\citep{wang-iccv21-pvt} & 81.0/71.1 & 44.5 & 66.0 & 48.3 & 40.7 & 63.4 & 43.7 & 43.4 & 63.6 & 46.1 & 26.1 & 46.0 & 59.5\\
Swin-S~\citep{liu-iccv21-swin} & 69.0/60.0 & 48.5 & 70.2 & 53.5 & 43.3 & 67.3 & 46.6 & 46.4 & 67.0 & 50.1 & 31.0 & 50.1 & 60.3\\
Swin-B~\citep{liu-iccv21-swin} & 107.0/98.0 & 48.5 & 69.8 & 53.2 & 43.4 & 66.8 & 49.6 & 45.8 & 66.4 & 49.1 & 29.9 & 49.4 & 60.3\\
Focal-Small~\citep{yang-21-focal} & 71.2/61.7 & 48.8 & 70.5 & 53.6 & 43.8 & 67.7 & 47.2& 47.3 & 67.8 & 51.0 & 31.6 & 50.9 & 61.1  \\
Focal-Base~\citep{yang-21-focal} & 110.0/100.8 & 49.0 & 70.1 & 53.6 & 43.7 & 67.6 & 47.0& 46.9 & 67.8 & 50.3 & 31.9 & 50.3 & 61.5 \\
Uniformer-B$_{h14}$~\citep{li-22-uniformer} & 69.0/-- & 50.3 & 72.7 & 55.3 & 44.8 & 69.0 & 48.3 & -- & -- & -- & -- & -- & --\\
\sota{\model-B} & \sota{65.6/55.5} & \sota{\textbf{51.5}} & \sota{72.7} & \sota{56.8} & \sota{\textbf{45.9}} & \sota{70.0} & \sota{50.0} 
 & \sota{\textbf{50.2}} & \sota{71.7} & \sota{55.3} & \sota{36.4} & \sota{52.1} & \sota{62.3} \\
\hline
\end{tabular}}
\label{tab:maskrcnn-3x}
\vspace{-1.5em}
\end{table*}

%% file: src/tables/exp-segmentation.tex
\begin{table}[!ht]
\renewcommand{\tabcolsep}{1.8mm}
\scriptsize
\caption{\textbf{Semantic segmentation on ADE20k~\citep{zhou-19ijcv-ade20k} with UperNet~\citep{xiao-18eccv-upernet} and Semantic FPN~\citep{kirillov-19cvpr-fpn}.} All the models are pre-trained on ImageNet-1K~\citep{russakovsky-15ijcv-imagenet} and finetuned with task-specific heads. We follow the standard training and evaluation processes in~\citep{mmseg2020} for fair comparisons.}
\vspace{-1.5em}
\begin{center}
\scriptsize
\begin{tabular}{l|cc|cc}
\hline
\multirow{2}{*}{Backbone}  & \multicolumn{2}{c|}{Semantic FPN} & \multicolumn{2}{c}{UperNet}\\
 & \#Param(M)  & mIoU(\%) & \#Param(M)  & mIoU(\%) \\
\hline
ResNet18~\citep{he-16cvpr-resnet}  & 15.5 & 32.9  & -- & --\\
PVT-Tiny~\citep{wang-iccv21-pvt}  & 17.0 & 35.7 & -- & -- \\
XCiT-T12/16~\citep{ali-21nips-xcit} & 8.4 & 38.1 & 33.7 & 41.5 \\
XCiT-T12/8~\citep{ali-21nips-xcit} & 8.4 & 39.9 & 33.7 & 43.5 \\
\sota{\model-T}  & \sota{14.1} & \sota{\textbf{46.2}} & \sota{39.1} & \sota{\textbf{47.4}} \\

\hline
ResNet50~\citep{he-16cvpr-resnet}  & 28.5 & 36.7 & 66.5 & 42.0 \\
PVT-Small~\citep{wang-iccv21-pvt}  & 28.2 & 39.8 & -- & -- \\
Swin-T~\citep{liu-iccv21-swin} & -- & -- & 59.9 & 44.5 \\
Focal-T~\citep{yang-21-focal}  & -- & -- & 62 & 45.8 \\
XCiT-S12/16~\citep{ali-21nips-xcit} & 30.4 & 43.9 & 52.4 & 45.9 \\
XCiT-S12/8~\citep{ali-21nips-xcit} & 30.4 & 44.2 & 52.4 & 46.6 \\
\sota{\model-S}  & \sota{26.3} & \sota{\textbf{47.8}}& \sota{51.1} & \sota{\textbf{49.6}} \\
\hline
ResNet101~\citep{he-16cvpr-resnet} & 47.5 & 38.8 & 85.5 & 43.8 \\
ResNeXt101-32x4d~\citep{xie-17cvpr-resnext} & 47.1 & 41.6  & -- & -- \\
ResNeXt101-64x4d~\citep{xie-17cvpr-resnext} & 65.1 & 44.8 & -- & -- \\
PVT-Large~\citep{wang-iccv21-pvt} & 65.1 & 42.1 & -- & -- \\
Swin-S~\citep{liu-iccv21-swin} & -- & -- & 81.0 & 47.6 \\
Swin-B~\citep{liu-iccv21-swin} & -- & -- & 121.0 & 48.1 \\
Focal-S~\citep{yang-21-focal}  & -- & -- & 85.0 & 50.0 \\
Focal-B~\citep{yang-21-focal}  & -- & -- & 126.0 & 50.5 \\
XCiT-M24/16~\citep{ali-21nips-xcit} & 90.8 & 45.9 & 109.0 & 47.6 \\
XCiT-M24/8~\citep{ali-21nips-xcit} & 90.8 & 46.9 & 109.0 & 48.4 \\
\sota{\model-B}  & \sota{49.7} & \sota{\textbf{50.0}} & \sota{75.5} & \sota{\textbf{51.2}} \\

\hline
\end{tabular}
\end{center}
\label{tab:ade20k}
\vspace{-4em}
\end{table}

%% file: src/tables/exp-abla-aggregators.tex
\begin{table*}[t!]
\centering
\renewcommand{\arraystretch}{0.9}
\renewcommand{\tabcolsep}{3.5mm}
\scriptsize
\caption{\textbf{Ablation studies on the group aggregators in \block\ Block.} We use $\operatorname{Agg}^1$, $\operatorname{Agg}^2$, and $\operatorname{Agg}^3$ to denote aggregators (in the pre-attention branch) with kernel sizes of 3, 5, and 7, respectively, and $\operatorname{Agg}^0$ to denote the aggregator (in the non-attention branch) as shown in ~\cref{fig:pipeline}. We report the Top-1 accuracy on ImageNet-1k together with $\mathrm{AP}^m$ and $\mathrm{AP}^b$ on COCO.}
\begin{tabular}{l|c|ccc|cc|ccc}
    \hline
	Method & $\operatorname{Agg}^{0}$  & $\operatorname{Agg}^{1}(3\times 3)$ & $\operatorname{Agg}^{2}(5\times 5)$ & $\operatorname{Agg}^{3}(7\times 7)$ & \#Params.(M) & \#GFLOPs & Top-1 (\%) & $\mathrm{AP}^b$ & $\mathrm{AP}^m$ \\

	\hline
	\model-T & \usym{2613} &  \usym{2613} & \usym{2613} & \usym{2613} & 10.5 & 3.4 & 80.9 & 43.9 & 39.6 \\
	\model-T & \usym{2613} &  \usym{1F5F8} & \usym{1F5F8} & \usym{1F5F8} & 10.8 & 3.5 & 81.9~\textcolor{blue}{(+1.0)} & 44.6 & 40.4 \\
	\model-T & \usym{1F5F8} &  \usym{2613} & \usym{2613} & \usym{2613} & 10.7 & 3.5 & 81.3~\textcolor{blue}{(+0.4)} & 44.1 & 40.3 \\
	\hline
	\model-T & \usym{1F5F8} &  \usym{1F5F8} & \usym{2613} & \usym{2613}  & 10.8 & 3.6 & 82.2~\textcolor{blue}{(+1.3)} & 44.6 & 40.3 \\
	\model-T & \usym{1F5F8} &  \usym{2613} & \usym{1F5F8} & \usym{2613}  & 10.8 & 3.6 & 82.3~\textcolor{blue}{(+1.4)} & 44.6 & 40.4\\
	\model-T & \usym{1F5F8} &  \usym{2613} & \usym{2613} & \usym{1F5F8}  & 10.9 & 3.6 & 82.3~\textcolor{blue}{(+1.4)} & 44.8 & 40.3 \\
	\model-T & \usym{1F5F8} &  \usym{1F5F8} & \usym{1F5F8} & \usym{1F5F8} & 11.0 & 3.7 & 82.5~\textcolor{blue}{(+1.6)} & 45.4 & 40.6 \\
\hline
\end{tabular}
\label{tab:abla_aggmodule}
\vspace{-1.5em}
\end{table*}

%% file: src/tables/exp-abla-aggregator-implementation.tex
\begin{table}[t!]
\renewcommand{\tabcolsep}{0.8mm}
\centering
\renewcommand{\arraystretch}{0.9}
\renewcommand{\tabcolsep}{3mm}
\scriptsize
\caption{ImageNet-1k classification, COCO det and instance seg (1x with Mask RCNN) performances on various aggregators.}
\vspace{-1em}
\begin{tabular}{l|c|c|cc}
    \hline
	Method & Implementation  & Top-1 (\%) & $\mathrm{AP}^b$ & $\mathrm{AP}^m$ \\
	\hline
	\model-T & MinPool  & 82.3 & 42.4 & 39.7 \\
	\model-T & MaxPool  & 82.2 & 42.3 & 39.7 \\
	\model-T & AvgPool  & 82.2 & 42.3 & 39.6 \\
	\model-T & DWConv   & 82.5 & 42.5 & 39.8 \\
\hline
\end{tabular}
\label{tab:abla_implementation}
\vspace{-0.5em}
\end{table}

%% file: src/tables/exp-abla-gma-necessary.tex
\begin{table}[t!]
\centering
\renewcommand{\arraystretch}{0.9}
\renewcommand{\tabcolsep}{4mm}
\scriptsize
\caption{ImageNet-1k validation of replacing GMA with other attention modules on \model-T.}
\vspace{-1em}
\begin{tabular}{l|c|cc}
    \hline
	Attention Type & \#Params.(M) & \#GFLOPs & Top-1 (\%)  \\
	\hline
	GMA & 10.5 & 3.4 & 82.5 \\
	Swin-attention & 10.8 & 3.5 & 79.9 \textcolor{blue}{(-2.6)} \\
	PVT-attention & 16.3 & 3.3 & 79.1 \textcolor{blue}{(-3.4)} \\
\hline
\end{tabular}
\label{tab:abla_marco}
\vspace{-1em}
\end{table}

%% file: src/tables/exp-abla-kernel-configuration.tex
\begin{table}[!ht]
\centering
\renewcommand{\tabcolsep}{1mm}
\renewcommand{\tabcolsep}{2mm}
\scriptsize
\caption{Explorations on optimal kernel configurations with GroupMixFormer-T}

\vspace{-1em}
    \begin{tabular}{l |c c c}
    \hline
     Strategy   & \#Params.(M) & \#GFLOPS & Top-1 Acc (\%)  \\
    \hline
        kernel sizes = [5,7,9] & 11.2 & 3.9 & 82.0 \\
        large kernel to small kernel & 10.8 & 3.7  & 82.2 \\
        small kernel to large kernel & 11.0 & 3.7 & 82.0 \\
        Ours (\model-T)  & 11.0 & 3.7 & \textbf{82.5} \\
    \hline
    \end{tabular} 
\label{tab:abla_kernel_configuration}
\vspace{-0.5em}
\end{table}

%% file: src/tables/exp-abla-gma-generalization.tex
\begin{table}[t!]
\centering
\renewcommand{\tabcolsep}{0.8mm}
\renewcommand{\tabcolsep}{2.8mm}
\scriptsize
\caption{ImageNet-1k validation accuracy of incorporating aggregators to other ViT architectures.}
\vspace{-1em}
\begin{tabular}{l|c|cc|c}
\hline
	Structures &  Aggregators & \#Params.(M) & \#GFLOPs & Top-1 (\%)  \\
	\hline
	Swin-T & \usym{2613}  & 28.3 & 4.5 & 81.3 \\
	Swin-T & \usym{1F5F8}  & 28.8 & 4.8 & 81.8~\textcolor{blue}{(+0.5)}\\
	\hline
	PVT-Small & \usym{2613}  & 24.5 & 3.8 & 79.8 \\
	PVT-Small & \usym{1F5F8}  & 25.2 & 4.0 & 80.6~\textcolor{blue}{(+0.8)}\\
\hline
\end{tabular}
\label{tab:abla_others}
\vspace{-1.5em}
\end{table}

%% file: src/secs/5-conclusion.tex
\section{Conclusion}
In this paper, we proposed an advanced attention mechanism, named Group-Mix Attention (GMA). In contrast to the popular multi-head self-attention (MHSA) that only models the correlations among individual tokens, the proposed GMA utilizes the group aggregators to simultaneously capture the token-to-token, token-to-group, and group-to-group correlations. We proposed \model\ based on GMA and instantiated a series of practical visual backbones with different sizes. Extensive experiments on the standard visual recognition benchmarks (including image classification, object detection, and semantic segmentation) have validated the effectiveness of the proposed GMA and \model.

%% file: src/secs/6-appendix.tex
\setcounter{section}{0}
\renewcommand{\thesection}{\Alph{section}}
\begin{center}
	\LARGE \text{Appendix}
\end{center}

This appendix includes detailed illustrations of the algorithm, training configurations and additional experiments. In ~\cref{alg:code}, we present the PyTorch-style pseudocode of GMA Block for easy implementation. In ~\cref{sec:detail}, we detail the attention computation and elaborate on the training configurations for image classification, object detection, and instance/semantic segmentation. Besides, in ~\cref{sec:appendix_B} and ~\cref{sec:appendix_C}, we present additional experiments and visualizations to further validate \model's effectiveness, respectively. 

\begin{algorithm*}[h]
\caption{PyTorch-style Pseudocode of \block\ Block.}
\label{alg:code}
\definecolor{codeblue}{rgb}{0.25,0.5,0.5}
\lstset{
  backgroundcolor=\color{white},
  basicstyle=\fontsize{8.8pt}{8.8pt}\ttfamily\selectfont,
  columns=fullflexible,
  breaklines=true,
  captionpos=b,
  commentstyle=\fontsize{7.2pt}{7.2pt}\color{codeblue},
  keywordstyle=\fontsize{7.2pt}{7.2pt},
}

\begin{lstlisting}[language=python]
# x: the input token with shape of (B, N, D), B is batch size, N=H*W, D is dimension
# qkv_mapping(): linear mapping (in=D, out=D*3) to generate Q, K, V
# att(): efficient multi-head Q-K-V computation
# token_ensemble(): linear mapping (in=out=D) to combine the outputs from the attention and non-attention branches
# act: activation function, implemented by HardSwish
# norm: normalization function, implemented by LayerNorm
# The aggregator is implemented by a depth-wise convolution (channels=groups=D//5) following a linear mapping
def GMA(x):
    B,N,D=x.shape
    split_dim = D//5
    
    # Generate Q/K/V
    qkv = qkv_mapping(x).reshape(B, N, 3, D).permute(2, 0, 1, 3).reshape(3*B, N, D)
    qkv = qkv.transpose(1, 2).view(3*B, D, H, W)
    qkv = qkv.split([split_dim]*5, dim=1)
    # Now qkv[i] is the i-th branch with shape of (3*B, split_dim, H, W)
    
    qkv_pre_att_0 = act(norm(qkv[0]))
    # Generate group proxies via different aggregators
    qkv_pre_att_1 = act(norm(aggregator_pre_att_3x3(qkv[1])))
    qkv_pre_att_2 = act(norm(aggregator_pre_att_5x5(qkv[2])))
    qkv_pre_att_3 = act(norm(aggregator_pre_att_7x7(qkv[3])))
    
    # Non-attention branch 
    qkv_non_att = qkv[4].reshape(3, B, split_dim, H, W).permute(1, 0, 2, 3, 4).reshape(B, 3*split_dim, H, W)
    x_non_att = act(norm(aggregator_non_att_3x3(qkv_non_att)).reshape(B, split_dim, H, W))
    
    # Efficient multi-head Q-K-V self-Attention. We ignore the number of heads for brevity
    # Its input is (3*B, D*4/5, H, W), output is (B, D*4/5, H, W)
    qkv_input = torch.cat([qkv_pre_att_0, qkv_pre_att_1, qkv_pre_att_2, qkv_pre_att_3], dim=1)
    x_att = att(qkv_input)
    
    # combine the outputs from attention and the non-attention branch 
    x = torch.cat([x_att, x_non_att], dim=1)    # the shape becomes (B, D, H, W)
    x = x.reshape(B, D, N).permute(0, 2, 1)     # the shape becomes (B, N, D)
    x = token_ensemble(x)
    return x
\end{lstlisting}
\end{algorithm*}

\section{Implementation Details} \label{sec:detail}
\subsection{Attention Computation}
We  detail the attention computation adopted by \model\ in this appendix. The Q/K/V entries are first uniformly divided into five segments where we perform group aggregation on four segments. We use $\mathrm{X}^q_i, \mathrm{X}^k_i, \mathrm{X}^v_i$ ($i \in [1,2,3,4]$) to denote the segments divided from Q/K/V entries, respectively. To produce the group proxies $Q^\prime$, $K^\prime$, and $V^\prime$, we first employ the aggregation operation on the segments as $\operatorname{Agg}^i(\mathrm{X}^q_i)$, $\operatorname{Agg}^i(\mathrm{X}^k_i)$ and $\operatorname{Agg}^i(\mathrm{X}^v_i)$. Then we concatenate all the four ($i \in [1,2,3,4]$) aggregated features  to output group proxies $Q^\prime$, $K^\prime$, and $V^\prime$. Afterward, we perform attention computation as introduced in ~\citep{xu-21iccv-coat,ali-21nips-xcit,shen-21IWCACV-efficient} on the group proxies to generate the final output $\operatorname{Att}$.
\vspace{-2mm}
\begin{equation}\label{eq:att}
    \operatorname{Att} = 
     \frac{Q^\prime}{\sqrt{d}}\operatorname{Softmax}({K^\prime}^{T}{V^\prime}).
\nonumber
\vspace{-2mm}
\end{equation}

\begin{table*}[t]
\centering
\renewcommand{\arraystretch}{0.9}
\renewcommand{\tabcolsep}{1mm}
\small
\caption{Comparisons on inference speed with different models.}
\begin{tabular}{l|ccccc}
    \hline
	Method  & Swin-T  & PVT-S & CSWin-T  & \model-S & \model-S(AvgPool) \\
    \hline
        Throughput (images/s) & 755 &  820 & 701 & 596 & 611 \\
        \#Param.(M) &29.0 &24.5 & 23.0& 22.4& 22.1 \\
        \#FLOPs.(G) &4.5 &3.8 & 4.3 & 5.2& 5.0 \\
        Performance (\%) &81.3 &79.8 & 82.7& 83.4& 83.0\\
    \hline
\end{tabular}
\label{tab:supp_speed}
\end{table*}

\begin{table*}[!ht]
\centering
\caption{\textbf{Object detection and instance segmentation performance} on COCO 2017~\citep{lin-14eccv-coco} with Cascade Mask R-CNN~\citep{cai-19pami-cascade}. 
}
\small
\begin{tabular}{l|c|cccccc}
\hline
\renewcommand{\arraystretch}{0.1}
\multirow{2}{*}{Backbone} &\multicolumn{7}{c}{Cascade Mask R-CNN 3$\times$ + MS}  \\
\cline{2-8} 
 & \#P (M) & $\mathrm{AP}^b$ & $\mathrm{AP}^b_{50}$ & $\mathrm{AP}^b_{75}$ & $\mathrm{AP}^m$ & $\mathrm{AP}^m_{50}$ & $\mathrm{AP}^m_{75}$ 
 \\
\hline
ResNet50~\citep{he-16cvpr-resnet} & 82.0 & 46.3 & 64.3 & 50.5 & -- & -- & -- \\
PVTv2-b2-Linear~\citep{wang-21cvm-pvtv2} & 80.1 & 50.9 & 69.5 & 55.2 & 44.0 & 66.8 & 47.7 \\
PVTv2-b2~\citep{wang-21cvm-pvtv2} & 82.9 & 51.1  & 69.8 & 55.3 & 44.4 & 67.2 & 48.1  \\
Swin-T~\citep{liu-iccv21-swin} & 85.6 & 50.2 & 68.8 & 54.7 & 43.5 &66.1 &46.9 \\
Focal-T~\citep{xu-21iccv-coat} & 86.7 & 51.5 & 70.6 & 55.9 & -- & -- & -- \\
\sota{\model-T (ours)} & \sota{68.6} & \sota{\textbf{51.5}} & \sota{70.2} & \sota{55.7} & \sota{\textbf{44.4}} & \sota{67.5} & \sota{48.2}  \\
\sota{\model-S (ours)} & \sota{80.0} & \sota{\textbf{51.9}} & \sota{70.7} & \sota{56.1} & \sota{\textbf{45.1}} & \sota{68.3} & \sota{48.4} \\
\hline
\end{tabular}
\label{tab:cascaded-rcnn}
\end{table*}

\renewcommand{\tabcolsep}{2pt}
\def\swfive{0.13\linewidth}
\begin{figure*}[t]
\footnotesize
\begin{center}
\begin{tabular}{ccccccc}
\vspace{-0.2mm}
\includegraphics[width=\swfive]{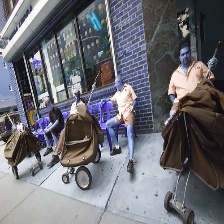}&
\includegraphics[width=\swfive]{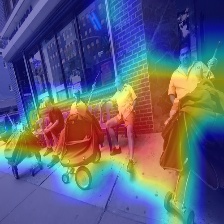}&
\includegraphics[width=\swfive]{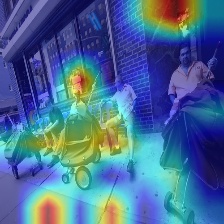}&
\includegraphics[width=\swfive]{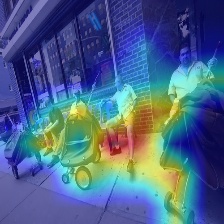}&
\includegraphics[width=\swfive]{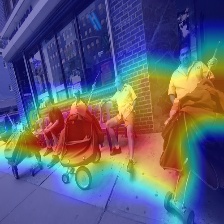}&
\includegraphics[width=\swfive]{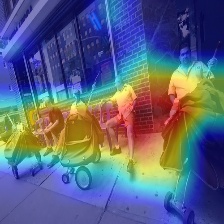}&
\includegraphics[width=\swfive]{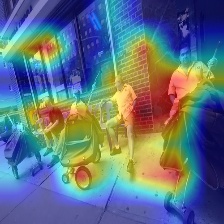}
\\

\includegraphics[width=\swfive]{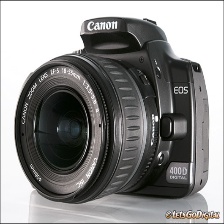}&
\includegraphics[width=\swfive]{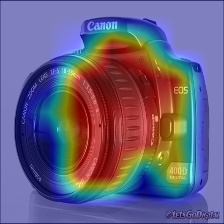}&
\includegraphics[width=\swfive]{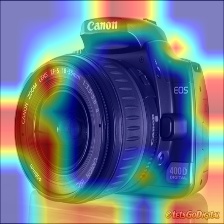}&
\includegraphics[width=\swfive]{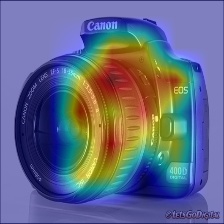}&
\includegraphics[width=\swfive]{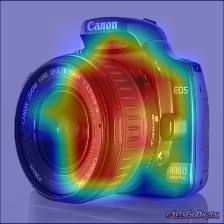}&
\includegraphics[width=\swfive]{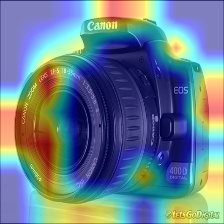}&
\includegraphics[width=\swfive]{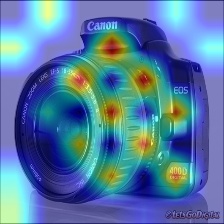}
\\

\includegraphics[width=\swfive]{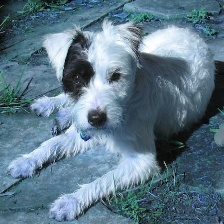}&
\includegraphics[width=\swfive]{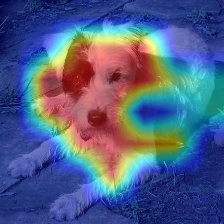}&
\includegraphics[width=\swfive]{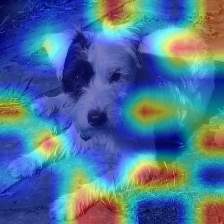}&
\includegraphics[width=\swfive]{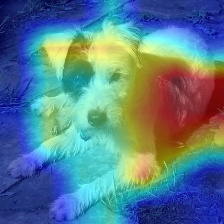}&
\includegraphics[width=\swfive]{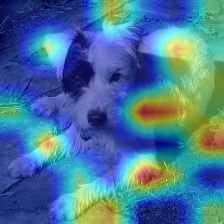}&
\includegraphics[width=\swfive]{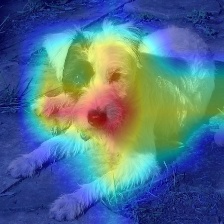}&
\includegraphics[width=\swfive]{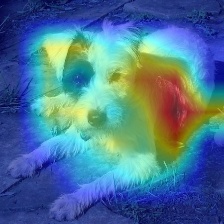}
\\

\includegraphics[width=\swfive]{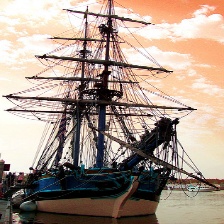}&
\includegraphics[width=\swfive]{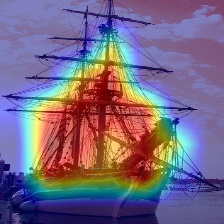}&
\includegraphics[width=\swfive]{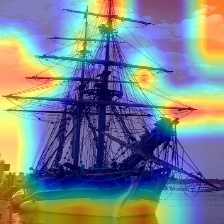}&
\includegraphics[width=\swfive]{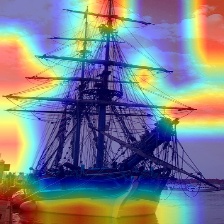}&
\includegraphics[width=\swfive]{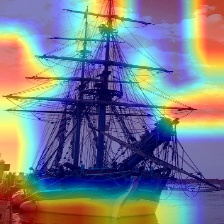}&
\includegraphics[width=\swfive]{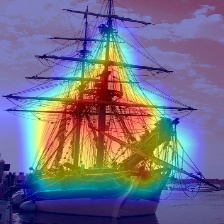}&
\includegraphics[width=\swfive]{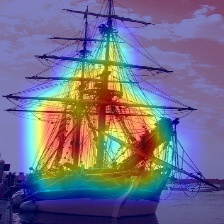}
\\

\includegraphics[width=\swfive]{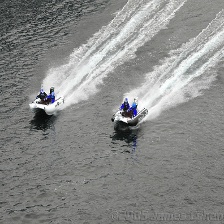}&
\includegraphics[width=\swfive]{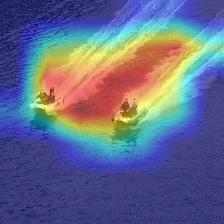}&
\includegraphics[width=\swfive]{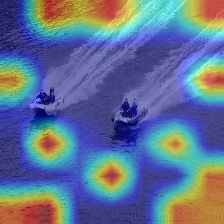}&
\includegraphics[width=\swfive]{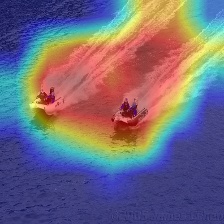}&
\includegraphics[width=\swfive]{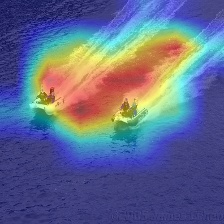}&
\includegraphics[width=\swfive]{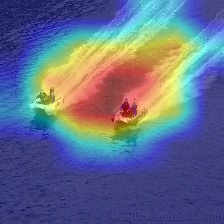}&
\includegraphics[width=\swfive]{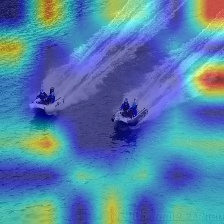}
\\

\includegraphics[width=\swfive]{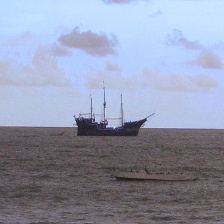}&
\includegraphics[width=\swfive]{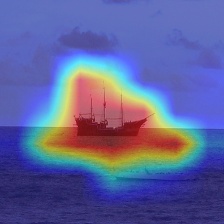}&
\includegraphics[width=\swfive]{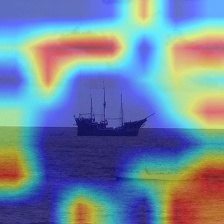}&
\includegraphics[width=\swfive]{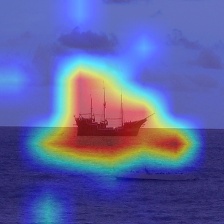}&
\includegraphics[width=\swfive]{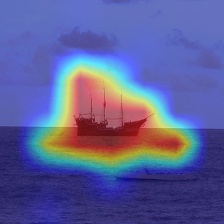}&
\includegraphics[width=\swfive]{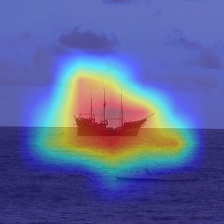}&
\includegraphics[width=\swfive]{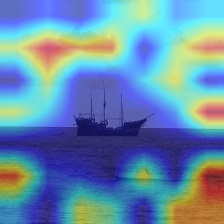}
\\

(a) Input & (b) All & (c) $1\times1$ & (d) $3\times3$ & (e) $5\times5$  & (f) $7\times7$ & (g) Non-attention\\
\end{tabular}
\end{center}
\caption{\textbf{Attention visualizations on \model-S.} The model attends to the pixels marked as red more than the others. Input images are shown in (a). In (c) to (f), we show the attention response maps from different aggregators in the pre-attention branches. In (g), we show the response maps from the aggregators in the non-attention branch. The combined response maps (outputs from the token ensemble layer) are shown in (b).}
\vspace{-1em}
\label{fig:vis}
\end{figure*}

\subsection{Image Classification} 
The standard ImageNet-1K dataset~\citep{russakovsky-15ijcv-imagenet} contains about 1.3 million training samples and 50K validation samples from 1000 categories. We experiment with input resolutions of $224\times 224$, $384\times 384$, or $448\times 448$. We follow~\citep{touvron-21ICML-deit} for data augmentation, including Mixup~\citep{zhang-17-mixup}, CutMix~\citep{yun-19iccv-cutmix}, random erasing~\citep{zhong-20aaai-random}, etc. We use the same training recipes as \citep{liu-iccv21-swin}. For training with $224\times 224$, all \model\ instances are trained for 300 epochs with a batch size of 1024. The initial learning rate is set to $\mathrm{10}^{-3}$ with a linear warm-up for 20 epochs and then cosine annealing towards zero. We adopt the AdamW optimizer~\citep{loshchilov-17-adamw} with a weight decay coefficient of 0.05. The drop-path rates~\citep{huang-16eccv-dpr} are set to 0.0, 0.1, 0.2, 0.4 and 0.5 for \model-M/T/S/B/L, respectively. Besides, for higher resolutions (i.e., $384\times 384$ and $448\times 448$), we finetune the $224\times 224$-pretrained models for another 30 epochs with an initial learning rate of $2\times\mathrm{10}^{-6}$ and a linear warm-up for 5 epochs and then cosine annealing. For finetuning, we use AdamW optimizer~\citep{loshchilov-17-adamw} with a weight decay coefficient of $1.0\times\mathrm{10}^{-8}$.

\subsection{Object Detection and Instance segmentation} \label{sec:implementation_det}
For object detection, we experiment on COCO 2017~\citep{lin-14eccv-coco} with Mask R-CNN~\citep{he-17iccv-maskrcnn} and RetinaNet~\citep{lin-17iccv-retinanet}. All models are initialized with the weights pretrained on ImageNet-1K~\citep{russakovsky-15ijcv-imagenet}. The detectors are finetuned on COCO train2017 (118k images) and evaluated on COCO val2017 (5k images). For data augmentation, we adopt multi-scale training as a common practice~\citep{liu-iccv21-swin}. We also follow the standard $3\times$ (36-epoch) training schedules provided in~\citep{mmdetection}. We use AdamW~\citep{loshchilov-17-adamw} with a weight decay coefficient of 0.05 for Mask R-CNN and $\mathrm{10}^{-4}$ for RetinaNet.

For instance segmentation, we benchmark \model\ models on COCO 2017~\citep{lin-14eccv-coco} with Mask R-CNN~\citep{he-17iccv-maskrcnn} with the same configurations as described above.

Moreover, we present additional results with Cascade Mask R-CNN~\citep{cai-19pami-cascade} in this supplementary material. We use the same training configurations as Mask R-CNN. 

\subsection{Semantic Segmentation} 

For semantic segmentation, we experiment on ADE20k~\citep{zhou-19ijcv-ade20k} with UperNet~\citep{xiao-18eccv-upernet} and Semantic FPN~\citep{kirillov-19cvpr-fpn}). ADE20K contains $\sim$20k, $\sim$2k, and $\sim$3k images for training, validation, and testing, respectively, from 150 categories. Following common practices~\citep{wang-iccv21-pvt,liu-iccv21-swin}, we randomly resize and crop the image to $512 \times 512$ for training, and rescale the shorter side to 512 pixels for testing. We use AdamW with a weight decay coefficient of $\mathrm{10^{-4}}$ for Semantic FPN and 0.01 for UperNet. The Semantic FPN is trained for 80k iterations while the UperNet is trained for 160k iterations. The learning rate is initialized as $6\times 10^{-5}$, warmed up linearly in 1500 iterations, and then decayed following the polynomial decay schedule with a power of 0.9.

\section{Speed Analysis} 
We empirically found that implementing the aggregators in GMA with DW-Conv indeed slow-down the inference speed. For instance, as shown in \cref{tab:supp_speed}, when tested on the single V100 GPU, our throughput (596 images/s) is smaller than the prevalent backbones (e.g., Swin-T with 755 images/s, CSWin-T with 701 images/s). However, our model outperforms others by large margins in recognition performance. Besides, it's noteworthy that with accuracy maintained, the speed of \model could be further improved by implementing with more efficient aggregators (e.g., +15 image/s by AvgPool as shown in \cref{tab:supp_speed}).

\section{Additional results with Cascade Mask R-CNN}   \label{sec:appendix_B}
To further verify the effectiveness of our proposed model, we equip \model\ with a more powerful object detector, i.e., Cascaded Mask R-CNN~\citep{cai-19pami-cascade}. Detailed implementations are presented in \cref{sec:implementation_det}. Results in \cref{tab:cascaded-rcnn} show \model\ consistently outperforms the prevalent Transformer-based backbones (e.g., PVT-V2~\citep{wang-21cvm-pvtv2} and Swin~\citep{liu-iccv21-swin}). Besides, with fewer parameters (68.6M \textit{v.s.} 86.7M), our \model-T obtains a comparable performance with Focal-T (around 51.5\% AP$^\mathit{b}$). Our \model-S achieves new state-of-the-art performance with an AP$^\mathit{b}$ of 51.9\%.

\section{Attention Visualization} \label{sec:appendix_C}

We present attention response maps in \cref{fig:vis}. We show input images in (a), and the attention response maps from the ensemble layer in (b). Besides, the response maps of the outputs from the pre-attention branches and non-attention branch are shown in (c) to (g), respectively. We observe that applying self-attention on individual tokens sometimes fails to attend to the object, as shown in (c). In such a case, calculating the correlations among the group proxies, which are generated by the aggregators, may help. For example, as shown in the third row, calculating correlations among the groups, which are processed by aggregators with kernel sizes of 3 and 7, succeed in focusing on the dog, while modeling the token-to-token correlations in (c) focuses more on the background. These results indicate that there exist some patterns so that some tokens should be handled as a whole to capture the object features. In \block, the representations captured by different aggregators are combined. It validates that comprehensively modeling the token-to-token, token-to-group, and group-to-group correlations leads to better vision recognition.

%% file: main.bbl
\begin{thebibliography}{69}
\providecommand{\natexlab}[1]{#1}
\providecommand{\url}[1]{\texttt{#1}}
\expandafter\ifx\csname urlstyle\endcsname\relax
  \providecommand{\doi}[1]{doi: #1}\else
  \providecommand{\doi}{doi: \begingroup \urlstyle{rm}\Url}\fi

\bibitem[Ali et~al.(2021)Ali, Touvron, Caron, Bojanowski, Douze, Joulin, Laptev, Neverova, Synnaeve, Verbeek, et~al.]{ali-21nips-xcit}
Alaaeldin Ali, Hugo Touvron, Mathilde Caron, Piotr Bojanowski, Matthijs Douze, Armand Joulin, Ivan Laptev, Natalia Neverova, Gabriel Synnaeve, Jakob Verbeek, et~al.
\newblock Xcit: Cross-covariance image transformers.
\newblock \emph{Advances in Neural Information Processing Systems}, 2021.

\bibitem[Ba et~al.(2016)Ba, Kiros, and Hinton]{ba-16-layer}
Jimmy~Lei Ba, Jamie~Ryan Kiros, and Geoffrey~E Hinton.
\newblock Layer normalization.
\newblock \emph{arXiv preprint arXiv:1607.06450}, 2016.

\bibitem[Bao et~al.(2021)Bao, Dong, and Wei]{bao-2021-beit}
Hangbo Bao, Li Dong, and Furu Wei.
\newblock Beit: Bert pre-training of image transformers.
\newblock \emph{arXiv preprint arXiv:2106.08254}, 2021.

\bibitem[Cai and Vasconcelos(2019)]{cai-19pami-cascade}
Zhaowei Cai and Nuno Vasconcelos.
\newblock Cascade r-cnn: high quality object detection and instance segmentation.
\newblock \emph{IEEE Transactions on Pattern Analysis and Machine Intelligence}, 2019.

\bibitem[Caron et~al.(2021)Caron, Touvron, Misra, J{\'e}gou, Mairal, Bojanowski, and Joulin]{caron-21iccv-dino}
Mathilde Caron, Hugo Touvron, Ishan Misra, Herv{\'e} J{\'e}gou, Julien Mairal, Piotr Bojanowski, and Armand Joulin.
\newblock Emerging properties in self-supervised vision transformers.
\newblock In \emph{IEEE/CVF International Conference on Computer Vision}, 2021.

\bibitem[Chen et~al.(2021{\natexlab{a}})Chen, Dao, Liang, Yang, Song, Rudra, and Re]{chen-21-butterfly}
Beidi Chen, Tri Dao, Kaizhao Liang, Jiaming Yang, Zhao Song, Atri Rudra, and Christopher Re.
\newblock Pixelated butterfly: Simple and efficient sparse training for neural network models.
\newblock \emph{arXiv preprint arXiv:2112.00029}, 2021{\natexlab{a}}.

\bibitem[Chen et~al.(2021{\natexlab{b}})Chen, Dao, Liang, Yang, Song, Rudra, and Re]{chen2021pixelated}
Beidi Chen, Tri Dao, Kaizhao Liang, Jiaming Yang, Zhao Song, Atri Rudra, and Christopher Re.
\newblock Pixelated butterfly: Simple and efficient sparse training for neural network models.
\newblock \emph{arXiv preprint arXiv:2112.00029}, 2021{\natexlab{b}}.

\bibitem[Chen et~al.(2021{\natexlab{c}})Chen, Dao, Winsor, Song, Rudra, and R{\'e}]{chen-21-scatterbrain}
Beidi Chen, Tri Dao, Eric Winsor, Zhao Song, Atri Rudra, and Christopher R{\'e}.
\newblock Scatterbrain: Unifying sparse and low-rank attention approximation.
\newblock \emph{arXiv preprint arXiv:2110.15343}, 2021{\natexlab{c}}.

\bibitem[Chen et~al.(2019)Chen, Wang, Pang, Cao, Xiong, Li, Sun, Feng, Liu, Xu, Zhang, Cheng, Zhu, Cheng, Zhao, Li, Lu, Zhu, Wu, Dai, Wang, Shi, Ouyang, Loy, and Lin]{mmdetection}
Kai Chen, Jiaqi Wang, Jiangmiao Pang, Yuhang Cao, Yu Xiong, Xiaoxiao Li, Shuyang Sun, Wansen Feng, Ziwei Liu, Jiarui Xu, Zheng Zhang, Dazhi Cheng, Chenchen Zhu, Tianheng Cheng, Qijie Zhao, Buyu Li, Xin Lu, Rui Zhu, Yue Wu, Jifeng Dai, Jingdong Wang, Jianping Shi, Wanli Ouyang, Chen~Change Loy, and Dahua Lin.
\newblock {MMDetection}: Open mmlab detection toolbox and benchmark.
\newblock \emph{arXiv preprint arXiv:1906.07155}, 2019.

\bibitem[Chen et~al.(2021{\natexlab{d}})Chen, Xie, and He]{chen-21iccv-mocov3}
Xinlei Chen, Saining Xie, and Kaiming He.
\newblock An empirical study of training self-supervised vision transformers.
\newblock In \emph{IEEE/CVF International Conference on Computer Vision}, 2021{\natexlab{d}}.

\bibitem[Chen et~al.(2022)Chen, Dai, Chen, Liu, Dong, Yuan, and Liu]{chen-22cvpr-mobileformer}
Yinpeng Chen, Xiyang Dai, Dongdong Chen, Mengchen Liu, Xiaoyi Dong, Lu Yuan, and Zicheng Liu.
\newblock Mobile-former: Bridging mobilenet and transformer.
\newblock In \emph{IEEE/CVF Conference on Computer Vision and Pattern Recognition}, 2022.

\bibitem[Chollet(2017)]{chollet-17cvpr-dwconv}
Fran{\c{c}}ois Chollet.
\newblock Xception: Deep learning with depthwise separable convolutions.
\newblock In \emph{IEEE/CVF Conference on Computer Vision and Pattern Recognition}, 2017.

\bibitem[Contributors(2020)]{mmseg2020}
MMSegmentation Contributors.
\newblock {MMSegmentation}: Openmmlab semantic segmentation toolbox and benchmark.
\newblock \url{https://github.com/open-mmlab/mmsegmentation}, 2020.

\bibitem[Cordonnier et~al.(2019)Cordonnier, Loukas, and Jaggi]{cordonnier-2019-relationship}
Jean-Baptiste Cordonnier, Andreas Loukas, and Martin Jaggi.
\newblock On the relationship between self-attention and convolutional layers.
\newblock \emph{arXiv preprint arXiv:1911.03584}, 2019.

\bibitem[Dai et~al.(2021)Dai, Chen, Xiao, Chen, Liu, Yuan, and Zhang]{dai-21cvpr-dynamic}
Xiyang Dai, Yinpeng Chen, Bin Xiao, Dongdong Chen, Mengchen Liu, Lu Yuan, and Lei Zhang.
\newblock Dynamic head: Unifying object detection heads with attentions.
\newblock In \emph{IEEE/CVF Conference on Computer Vision and Pattern Recognition}, pages 7373--7382, 2021.

\bibitem[Ding et~al.(2022)Ding, Xiao, Codella, Luo, Wang, and Yuan]{ding2022davit}
Mingyu Ding, Bin Xiao, Noel Codella, Ping Luo, Jingdong Wang, and Lu Yuan.
\newblock Davit: Dual attention vision transformers.
\newblock In \emph{European Conference on Computer Vision}, 2022.

\bibitem[Dong et~al.(2021)Dong, Bao, Chen, Zhang, Yu, Yuan, Chen, and Guo]{dong-2021-cswin}
Xiaoyi Dong, Jianmin Bao, Dongdong Chen, Weiming Zhang, Nenghai Yu, Lu Yuan, Dong Chen, and Baining Guo.
\newblock Cswin transformer: A general vision transformer backbone with cross-shaped windows.
\newblock \emph{arXiv preprint arXiv:2107.00652}, 2021.

\bibitem[Dosovitskiy et~al.(2021)Dosovitskiy, Beyer, Kolesnikov, Weissenborn, Zhai, Unterthiner, Dehghani, Minderer, Heigold, Gelly, et~al.]{dosovitskiy-iclr21-vit}
Alexey Dosovitskiy, Lucas Beyer, Alexander Kolesnikov, Dirk Weissenborn, Xiaohua Zhai, Thomas Unterthiner, Mostafa Dehghani, Matthias Minderer, Georg Heigold, Sylvain Gelly, et~al.
\newblock An image is worth 16x16 words: Transformers for image recognition at scale.
\newblock In \emph{International Conference on Learning Representations}, 2021.

\bibitem[Fan et~al.(2021)Fan, Xiong, Mangalam, Li, Yan, Malik, and Feichtenhofer]{fan-21iccv-multiscale}
Haoqi Fan, Bo Xiong, Karttikeya Mangalam, Yanghao Li, Zhicheng Yan, Jitendra Malik, and Christoph Feichtenhofer.
\newblock Multiscale vision transformers.
\newblock In \emph{IEEE/CVF International Conference on Computer Vision}, 2021.

\bibitem[Foret et~al.(2020)Foret, Kleiner, Mobahi, and Neyshabur]{foret-2020-sam}
Pierre Foret, Ariel Kleiner, Hossein Mobahi, and Behnam Neyshabur.
\newblock Sharpness-aware minimization for efficiently improving generalization.
\newblock \emph{arXiv preprint arXiv:2010.01412}, 2020.

\bibitem[Freeman et~al.(2018)Freeman, Roese-Koerner, and Kummert]{freeman2018effnet}
Ido Freeman, Lutz Roese-Koerner, and Anton Kummert.
\newblock Effnet: An efficient structure for convolutional neural networks.
\newblock In \emph{2018 25th ieee international conference on image processing (icip)}, 2018.

\bibitem[Graham et~al.(2021)Graham, El-Nouby, Touvron, Stock, Joulin, J{\'e}gou, and Douze]{graham-21iccv-levit}
Benjamin Graham, Alaaeldin El-Nouby, Hugo Touvron, Pierre Stock, Armand Joulin, Herv{\'e} J{\'e}gou, and Matthijs Douze.
\newblock Levit: a vision transformer in convnet's clothing for faster inference.
\newblock In \emph{IEEE/CVF International Conference on Computer Vision}, 2021.

\bibitem[He et~al.(2016)He, Zhang, Ren, and Sun]{he-16cvpr-resnet}
Kaiming He, Xiangyu Zhang, Shaoqing Ren, and Jian Sun.
\newblock Deep residual learning for image recognition.
\newblock In \emph{IEEE/CVF Conference on Computer Vision and Pattern Recognition}, 2016.

\bibitem[He et~al.(2017)He, Gkioxari, Doll{\'a}r, and Girshick]{he-17iccv-maskrcnn}
Kaiming He, Georgia Gkioxari, Piotr Doll{\'a}r, and Ross Girshick.
\newblock Mask r-cnn.
\newblock In \emph{IEEE/CVF International Conference on Computer Vision}, 2017.

\bibitem[Huang et~al.(2016)Huang, Sun, Liu, Sedra, and Weinberger]{huang-16eccv-dpr}
Gao Huang, Yu Sun, Zhuang Liu, Daniel Sedra, and Kilian~Q Weinberger.
\newblock Deep networks with stochastic depth.
\newblock In \emph{European Conference on Computer Vision}, 2016.

\bibitem[Huang et~al.(2017)Huang, Liu, Van Der~Maaten, and Weinberger]{huang-17cvpr-densenet}
Gao Huang, Zhuang Liu, Laurens Van Der~Maaten, and Kilian~Q Weinberger.
\newblock Densely connected convolutional networks.
\newblock In \emph{IEEE/CVF Conference on Computer Vision and Pattern Recognition}, 2017.

\bibitem[Jia et~al.(2021)Jia, Yang, Xia, Chen, Parekh, Pham, Le, Sung, Li, and Duerig]{jia-2021-clip}
Chao Jia, Yinfei Yang, Ye Xia, Yi-Ting Chen, Zarana Parekh, Hieu Pham, Quoc~V Le, Yunhsuan Sung, Zhen Li, and Tom Duerig.
\newblock Scaling up visual and vision-language representation learning with noisy text supervision.
\newblock \emph{arXiv preprint arXiv:2102.05918}, 2021.

\bibitem[Jiang et~al.(2021)Jiang, Hou, Yuan, Zhou, Shi, Jin, Wang, and Feng]{jiang-21nips-tokenlabeling}
Zi-Hang Jiang, Qibin Hou, Li Yuan, Daquan Zhou, Yujun Shi, Xiaojie Jin, Anran Wang, and Jiashi Feng.
\newblock All tokens matter: Token labeling for training better vision transformers.
\newblock \emph{Advances in Neural Information Processing Systems}, 34:\penalty0 18590--18602, 2021.

\bibitem[Kirillov et~al.(2019)Kirillov, Girshick, He, and Doll{\'a}r]{kirillov-19cvpr-fpn}
Alexander Kirillov, Ross Girshick, Kaiming He, and Piotr Doll{\'a}r.
\newblock Panoptic feature pyramid networks.
\newblock In \emph{IEEE/CVF Conference on Computer Vision and Pattern Recognition}, 2019.

\bibitem[Lee et~al.(2021)Lee, Kim, Willette, and Hwang]{lee-21-mpvit}
Youngwan Lee, Jonghee Kim, Jeff Willette, and Sung~Ju Hwang.
\newblock Mpvit: Multi-path vision transformer for dense prediction.
\newblock \emph{arXiv preprint arXiv:2112.11010}, 2021.

\bibitem[Li et~al.(2022{\natexlab{a}})Li, Wang, Zhang, Gao, Song, Liu, Li, and Qiao]{li-22-uniformer}
Kunchang Li, Yali Wang, Junhao Zhang, Peng Gao, Guanglu Song, Yu Liu, Hongsheng Li, and Yu Qiao.
\newblock Uniformer: Unifying convolution and self-attention for visual recognition.
\newblock \emph{arXiv preprint arXiv:2201.09450}, 2022{\natexlab{a}}.

\bibitem[Li et~al.(2022{\natexlab{b}})Li, Wu, Fan, Mangalam, Xiong, Malik, and Feichtenhofer]{li2022mvitv2}
Yanghao Li, Chao-Yuan Wu, Haoqi Fan, Karttikeya Mangalam, Bo Xiong, Jitendra Malik, and Christoph Feichtenhofer.
\newblock Mvitv2: Improved multiscale vision transformers for classification and detection.
\newblock In \emph{IEEE/CVF Conference on Computer Vision and Pattern Recognition}, 2022{\natexlab{b}}.

\bibitem[Liang et~al.(2022)Liang, Ge, Tong, Song, Wang, and Xie]{liang-22iclr-evit}
Youwei Liang, Chongjian Ge, Zhan Tong, Yibing Song, Jue Wang, and Pengtao Xie.
\newblock Not all patches are what you need: Expediting vision transformers via token reorganizations.
\newblock In \emph{International Conference on Learning Representations}, 2022.

\bibitem[Lin et~al.(2014)Lin, Maire, Belongie, Hays, Perona, Ramanan, Doll{\'a}r, and Zitnick]{lin-14eccv-coco}
Tsung-Yi Lin, Michael Maire, Serge Belongie, James Hays, Pietro Perona, Deva Ramanan, Piotr Doll{\'a}r, and C~Lawrence Zitnick.
\newblock Microsoft coco: Common objects in context.
\newblock In \emph{European Conference on Computer Vision}, 2014.

\bibitem[Lin et~al.(2017)Lin, Goyal, Girshick, He, and Doll{\'a}r]{lin-17iccv-retinanet}
Tsung-Yi Lin, Priya Goyal, Ross Girshick, Kaiming He, and Piotr Doll{\'a}r.
\newblock Focal loss for dense object detection.
\newblock In \emph{IEEE/CVF International Conference on Computer Vision}, 2017.

\bibitem[Liu et~al.(2021{\natexlab{a}})Liu, Hu, Lin, Yao, Xie, Wei, Ning, Cao, Zhang, Dong, et~al.]{liu-2021-swinv2}
Ze Liu, Han Hu, Yutong Lin, Zhuliang Yao, Zhenda Xie, Yixuan Wei, Jia Ning, Yue Cao, Zheng Zhang, Li Dong, et~al.
\newblock Swin transformer v2: Scaling up capacity and resolution.
\newblock \emph{arXiv preprint arXiv:2111.09883}, 2021{\natexlab{a}}.

\bibitem[Liu et~al.(2021{\natexlab{b}})Liu, Lin, Cao, Hu, Wei, Zhang, Lin, and Guo]{liu-iccv21-swin}
Ze Liu, Yutong Lin, Yue Cao, Han Hu, Yixuan Wei, Zheng Zhang, Stephen Lin, and Baining Guo.
\newblock Swin transformer: Hierarchical vision transformer using shifted windows.
\newblock In \emph{IEEE/CVF International Conference on Computer Vision}, 2021{\natexlab{b}}.

\bibitem[Liu et~al.(2022)Liu, Mao, Wu, Feichtenhofer, Darrell, and Xie]{liu-22-convnet}
Zhuang Liu, Hanzi Mao, Chao-Yuan Wu, Christoph Feichtenhofer, Trevor Darrell, and Saining Xie.
\newblock A convnet for the 2020s.
\newblock \emph{arXiv preprint arXiv:2201.03545}, 2022.

\bibitem[Loshchilov and Hutter(2017)]{loshchilov-17-adamw}
Ilya Loshchilov and Frank Hutter.
\newblock Decoupled weight decay regularization.
\newblock \emph{arXiv preprint arXiv:1711.05101}, 2017.

\bibitem[Ma et~al.(2018)Ma, Zhang, Zheng, and Sun]{ma-eccv18-shufflenetv2}
Ningning Ma, Xiangyu Zhang, Hai-Tao Zheng, and Jian Sun.
\newblock Shufflenet v2: Practical guidelines for efficient cnn architecture design.
\newblock In \emph{European Conference on Computer Vision}, 2018.

\bibitem[Mehta and Rastegari(2021)]{mehta-2021-mobilevit}
Sachin Mehta and Mohammad Rastegari.
\newblock Mobilevit: light-weight, general-purpose, and mobile-friendly vision transformer.
\newblock \emph{arXiv preprint arXiv:2110.02178}, 2021.

\bibitem[Paul and Chen(2021)]{paul-2021-robust}
Sayak Paul and Pin-Yu Chen.
\newblock Vision transformers are robust learners.
\newblock \emph{arXiv preprint arXiv:2105.07581}, 2021.

\bibitem[Raghu et~al.(2021)Raghu, Unterthiner, Kornblith, Zhang, and Dosovitskiy]{raghu-2021-vision}
Maithra Raghu, Thomas Unterthiner, Simon Kornblith, Chiyuan Zhang, and Alexey Dosovitskiy.
\newblock Do vision transformers see like convolutional neural networks?
\newblock \emph{arXiv preprint arXiv:2108.08810}, 2021.

\bibitem[Rao et~al.(2021)Rao, Zhao, Liu, Lu, Zhou, and Hsieh]{rao-21nips-dynamicvit}
Yongming Rao, Wenliang Zhao, Benlin Liu, Jiwen Lu, Jie Zhou, and Cho-Jui Hsieh.
\newblock Dynamicvit: Efficient vision transformers with dynamic token sparsification.
\newblock \emph{Advances in Neural Information Processing Systems}, 2021.

\bibitem[Ren et~al.(2022)Ren, Zhou, He, Feng, and Wang]{ren2022shunted}
Sucheng Ren, Daquan Zhou, Shengfeng He, Jiashi Feng, and Xinchao Wang.
\newblock Shunted self-attention via multi-scale token aggregation.
\newblock In \emph{IEEE/CVF Conference on Computer Vision and Pattern Recognition}, 2022.

\bibitem[Russakovsky et~al.(2015)Russakovsky, Deng, Su, Krause, Satheesh, Ma, Huang, Karpathy, Khosla, Bernstein, et~al.]{russakovsky-15ijcv-imagenet}
Olga Russakovsky, Jia Deng, Hao Su, Jonathan Krause, Sanjeev Satheesh, Sean Ma, Zhiheng Huang, Andrej Karpathy, Aditya Khosla, Michael Bernstein, et~al.
\newblock Imagenet large scale visual recognition challenge.
\newblock \emph{International Journal of Computer Vision}, 2015.

\bibitem[Shen et~al.(2021)Shen, Zhang, Zhao, Yi, and Li]{shen-21IWCACV-efficient}
Zhuoran Shen, Mingyuan Zhang, Haiyu Zhao, Shuai Yi, and Hongsheng Li.
\newblock Efficient attention: Attention with linear complexities.
\newblock In \emph{Proceedings of the IEEE/CVF Winter Conference on Applications of Computer Vision}, 2021.

\bibitem[Si et~al.(2022)Si, Yu, Zhou, Zhou, Wang, and Yan]{si2022inception}
Chenyang Si, Weihao Yu, Pan Zhou, Yichen Zhou, Xinchao Wang, and Shuicheng Yan.
\newblock Inception transformer.
\newblock \emph{Advances in Neural Information Processing Systems}, 2022.

\bibitem[Touvron et~al.(2021)Touvron, Cord, Douze, Massa, Sablayrolles, and J{\'e}gou]{touvron-21ICML-deit}
Hugo Touvron, Matthieu Cord, Matthijs Douze, Francisco Massa, Alexandre Sablayrolles, and Herv{\'e} J{\'e}gou.
\newblock Training data-efficient image transformers \& distillation through attention.
\newblock In \emph{International Conference on Machine Learning}, 2021.

\bibitem[Tu et~al.(2022)Tu, Talebi, Zhang, Yang, Milanfar, Bovik, and Li]{tu2022maxvit}
Zhengzhong Tu, Hossein Talebi, Han Zhang, Feng Yang, Peyman Milanfar, Alan Bovik, and Yinxiao Li.
\newblock Maxvit: Multi-axis vision transformer.
\newblock In \emph{European Conference on Computer Vision}, 2022.

\bibitem[Vaswani et~al.(2017)Vaswani, Shazeer, Parmar, Uszkoreit, Jones, Gomez, Kaiser, and Polosukhin]{vaswani-nips17-trans}
Ashish Vaswani, Noam Shazeer, Niki Parmar, Jakob Uszkoreit, Llion Jones, Aidan~N Gomez, {\L}ukasz Kaiser, and Illia Polosukhin.
\newblock Attention is all you need.
\newblock \emph{Advances in Neural Information Processing Systems}, 2017.

\bibitem[Wang et~al.(2018)Wang, Yao, Chen, Lin, Cai, He, and Liu]{wang2108crossformer}
W Wang, L Yao, L Chen, B Lin, D Cai, X He, and W Liu.
\newblock Crossformer: A versatile vision transformer hinging on cross-scale attention. arxiv 2021.
\newblock \emph{arXiv preprint arXiv:2108.00154}, 2018.

\bibitem[Wang et~al.(2021)Wang, Xie, Li, Fan, Song, Liang, Lu, Luo, and Shao]{wang-iccv21-pvt}
Wenhai Wang, Enze Xie, Xiang Li, Deng-Ping Fan, Kaitao Song, Ding Liang, Tong Lu, Ping Luo, and Ling Shao.
\newblock Pyramid vision transformer: A versatile backbone for dense prediction without convolutions.
\newblock In \emph{IEEE/CVF International Conference on Computer Vision}, 2021.

\bibitem[Wang et~al.(2022)Wang, Xie, Li, Fan, Song, Liang, Lu, Luo, and Shao]{wang-21cvm-pvtv2}
Wenhai Wang, Enze Xie, Xiang Li, Deng-Ping Fan, Kaitao Song, Ding Liang, Tong Lu, Ping Luo, and Ling Shao.
\newblock Pvtv2: Improved baselines with pyramid vision transformer.
\newblock \emph{Computational Visual Media}, 2022.

\bibitem[Wu et~al.(2021)Wu, Xiao, Codella, Liu, Dai, Yuan, and Zhang]{wu2021cvt}
Haiping Wu, Bin Xiao, Noel Codella, Mengchen Liu, Xiyang Dai, Lu Yuan, and Lei Zhang.
\newblock Cvt: Introducing convolutions to vision transformers.
\newblock In \emph{Proceedings of the IEEE/CVF International Conference on Computer Vision}, pages 22--31, 2021.

\bibitem[Wu et~al.(2022)Wu, Liu, Zhan, and Cheng]{wu2022p2t}
Yu-Huan Wu, Yun Liu, Xin Zhan, and Ming-Ming Cheng.
\newblock P2t: Pyramid pooling transformer for scene understanding.
\newblock \emph{IEEE Transactions on Pattern Analysis and Machine Intelligence}, 2022.

\bibitem[Xiao et~al.(2018)Xiao, Liu, Zhou, Jiang, and Sun]{xiao-18eccv-upernet}
Tete Xiao, Yingcheng Liu, Bolei Zhou, Yuning Jiang, and Jian Sun.
\newblock Unified perceptual parsing for scene understanding.
\newblock In \emph{European Conference on Computer Vision}, 2018.

\bibitem[Xie et~al.(2021{\natexlab{a}})Xie, Wang, Yu, Anandkumar, Alvarez, and Luo]{xie-21nips-segformer}
Enze Xie, Wenhai Wang, Zhiding Yu, Anima Anandkumar, Jose~M Alvarez, and Ping Luo.
\newblock Segformer: Simple and efficient design for semantic segmentation with transformers.
\newblock \emph{Advances in Neural Information Processing Systems}, 2021{\natexlab{a}}.

\bibitem[Xie et~al.(2017)Xie, Girshick, Doll{\'a}r, Tu, and He]{xie-17cvpr-resnext}
Saining Xie, Ross Girshick, Piotr Doll{\'a}r, Zhuowen Tu, and Kaiming He.
\newblock Aggregated residual transformations for deep neural networks.
\newblock In \emph{IEEE/CVF Conference on Computer Vision and Pattern Recognition}, 2017.

\bibitem[Xie et~al.(2021{\natexlab{b}})Xie, Lin, Yao, Zhang, Dai, Cao, and Hu]{xie-2021-swinself}
Zhenda Xie, Yutong Lin, Zhuliang Yao, Zheng Zhang, Qi Dai, Yue Cao, and Han Hu.
\newblock Self-supervised learning with swin transformers.
\newblock \emph{arXiv preprint arXiv:2105.04553}, 2021{\natexlab{b}}.

\bibitem[Xu et~al.(2021)Xu, Xu, Chang, and Tu]{xu-21iccv-coat}
Weijian Xu, Yifan Xu, Tyler Chang, and Zhuowen Tu.
\newblock Co-scale conv-attentional image transformers.
\newblock In \emph{IEEE/CVF International Conference on Computer Vision}, 2021.

\bibitem[Yang et~al.(2021)Yang, Li, Zhang, Dai, Xiao, Yuan, and Gao]{yang-21-focal}
Jianwei Yang, Chunyuan Li, Pengchuan Zhang, Xiyang Dai, Bin Xiao, Lu Yuan, and Jianfeng Gao.
\newblock Focal self-attention for local-global interactions in vision transformers.
\newblock \emph{arXiv preprint arXiv:2107.00641}, 2021.

\bibitem[Yuan et~al.(2021)Yuan, Hou, Jiang, Feng, and Yan]{yuan2021volo}
Li Yuan, Qibin Hou, Zihang Jiang, Jiashi Feng, and Shuicheng Yan.
\newblock Volo: Vision outlooker for visual recognition.
\newblock \emph{arXiv preprint arXiv:2106.13112}, 2021.

\bibitem[Yun et~al.(2019)Yun, Han, Oh, Chun, Choe, and Yoo]{yun-19iccv-cutmix}
Sangdoo Yun, Dongyoon Han, Seong~Joon Oh, Sanghyuk Chun, Junsuk Choe, and Youngjoon Yoo.
\newblock Cutmix: Regularization strategy to train strong classifiers with localizable features.
\newblock In \emph{IEEE/CVF International Conference on Computer Vision}, 2019.

\bibitem[Zhang et~al.(2017)Zhang, Cisse, Dauphin, and Lopez-Paz]{zhang-17-mixup}
Hongyi Zhang, Moustapha Cisse, Yann~N Dauphin, and David Lopez-Paz.
\newblock mixup: Beyond empirical risk minimization.
\newblock \emph{arXiv preprint arXiv:1710.09412}, 2017.

\bibitem[Zhang et~al.(2021)Zhang, Dai, Yang, Xiao, Yuan, Zhang, and Gao]{zhang-21iccv-visionlongformer}
Pengchuan Zhang, Xiyang Dai, Jianwei Yang, Bin Xiao, Lu Yuan, Lei Zhang, and Jianfeng Gao.
\newblock Multi-scale vision longformer: A new vision transformer for high-resolution image encoding.
\newblock In \emph{IEEE/CVF International Conference on Computer Vision}, 2021.

\bibitem[Zhong et~al.(2020)Zhong, Zheng, Kang, Li, and Yang]{zhong-20aaai-random}
Zhun Zhong, Liang Zheng, Guoliang Kang, Shaozi Li, and Yi Yang.
\newblock Random erasing data augmentation.
\newblock In \emph{Association for the Advancement of Artificial Intelligence}, 2020.

\bibitem[Zhou et~al.(2019)Zhou, Zhao, Puig, Xiao, Fidler, Barriuso, and Torralba]{zhou-19ijcv-ade20k}
Bolei Zhou, Hang Zhao, Xavier Puig, Tete Xiao, Sanja Fidler, Adela Barriuso, and Antonio Torralba.
\newblock Semantic understanding of scenes through the ade20k dataset.
\newblock \emph{International Journal of Computer Vision}, 2019.

\bibitem[Zhu et~al.(2023)Zhu, Wang, Ke, Zhang, and Lau]{zhu2023biformer}
Lei Zhu, Xinjiang Wang, Zhanghan Ke, Wayne Zhang, and Rynson~WH Lau.
\newblock Biformer: Vision transformer with bi-level routing attention.
\newblock In \emph{IEEE/CVF Conference on Computer Vision and Pattern Recognition}, 2023.

\end{thebibliography}
